\documentclass[review,sort&compress]{elsarticle}

\usepackage{float}

\usepackage{algorithm}
\usepackage{algorithmic}

\usepackage{amssymb}
\usepackage{latexsym}

\usepackage{url}

\usepackage{stfloats}

\usepackage{booktabs}

\usepackage{amsmath}
\usepackage{amsfonts}

\usepackage{hyperref} 
\hypersetup{
    colorlinks=true
}

\usepackage[labelfont=bf]{caption}

\usepackage{multirow}

\usepackage{tabularx} 

\usepackage[switch]{lineno}

\usepackage{xcolor}

\usepackage[british]{babel}

\usepackage{cleveref}
\crefname{subsection}{subsection}{subsections}

\usepackage{xstring}
\makeatletter
\AtBeginDocument{
\let\oldref\ref
\renewcommand{\ref}[1]{\IfBeginWith{#1}{fig:}%
{{\color{blue}Figure~\oldref{#1}}}%
{\IfBeginWith{#1}{tab:}{{\color{blue}Table~\oldref{#1}}}{Unsupported ref start}}}%
}

\usepackage{lipsum}
\makeatletter
\def\ps@pprintTitle{%
 \let\@oddhead\@empty
 \let\@evenhead\@empty
 \def\@oddfoot{}%
 \let\@evenfoot\@oddfoot}
\makeatother

\begin{document}

\begin{frontmatter}

\title{Detection of Face Recognition Adversarial Attacks}

\author[address]{Fabio Valerio Massoli\corref{cor1}}
\ead{fabio.massoli@isti.cnr.it}

\author[address]{Fabio Carrara}
\author[address]{Giuseppe Amato}
\author[address]{Fabrizio Falchi}

\cortext[cor1]{Corresponding author}
\address[address]{ISTI-CNR, via G. Moruzzi 1, 56124 Pisa, Italy}

\begin{abstract}
Deep Learning methods have become state-of-the-art for solving tasks such as Face Recognition (FR).
Unfortunately, despite their success, it has been pointed out that these learning models are exposed to \emph{adversarial} inputs --- images to which an imperceptible amount of noise for humans is added to maliciously fool a neural network --- thus limiting their adoption in sensitive real-world applications. 
While it is true that an enormous effort has been spent in order to train robust models against this type of threat, adversarial detection techniques have recently started to draw attention within the scientific community. A detection approach has the advantage that it does not require to re-train any model, thus it can be added on top of any system.
In this context, we present our work on adversarial samples detection in forensics mainly focused on detecting attacks against FR systems in which the learning model is typically used only as a features extractor. 
Thus, in these cases, train a more robust classifier might not be enough to defence a FR system.

In this frame, the contribution of our work is four-fold:
i) we tested our recently proposed adversarial detection approach against classifier attacks, i.e. adversarial samples crafted to fool a FR neural network acting as a classifier;
ii) using a k-Nearest Neighbor (kNN) algorithm as a guidance, we generated deep features attacks against a FR system based on a DL model acting as features extractor, followed by a kNN which gives back the query identity based on features similarity;
iii) we used the deep features attacks to fool a FR system on the 1:1 Face Verification task and we showed their superior effectiveness with respect to classifier attacks in fooling such type of system;
iv) we used the detectors trained on classifier attacks to detect deep features attacks, thus showing that such approach is generalizable to different types of offensives.

\end{abstract}

\begin{keyword}
Deep Learning \sep Face Recognition \sep Adversarial Attacks \sep Adversarial Detection \sep Adversarial Biometrics
\end{keyword}

\end{frontmatter}

\section{Introduction} \label{introduction}

Deep Learning (DL) quickly occupied a central role in recent AI-related technological breakthroughs covering multiple fields and applications: vision (e.g., image classification \citep{NIPS2012_4824}, object detection \citep{girshick2015fast}), natural language processing \citep{deng2018deep} and the combination of them (e.g.,  multi-modal \citep{Carrara2018}, sentiment analysis \citep{OrtisFB19}).
Despite achieving state-of-the-art performance in many scenarios, deep learning models still suffer from deficiencies that strongly limit their adoption in sensitive applications. Among others, the vulnerability of DL models in adversarial settings still poses challenges: it is relatively easy for an attacker to manipulate the output of a model by tampering its input often in an imperceptible way.
The existence of these perturbed inputs --- known as \emph{adversarial examples}~\citep{biggio2013evasion,szegedy2013intriguing} --- constitutes one of the major roadblocks in security-related applications such as DL-based biometrics systems for surveillance and access control that, despite performing brilliantly in natural settings~\citep{sundararajan2018deep}, can be easily evaded by knowledgeable adversaries.
Face Recognition enabled by Deep Neural Networks (DNN) is a case in point.
Several successful applications of deep models to FR have been proposed in the literature~\citep{cao2018vggface2,amato2018facial,liu2017sphereface}. Indeed, this kind of technology enables AI surveillance programs in multiple countries~\citep{feldstein2019global} and has already found its way into consumers products~\citep{sundararajan2018deep}.
However, researchers already showed how this kind of systems can be jeopardized by adversarial attacks both in the digital~\citep{dong2019efficient,song2018attacks} and physical domain~\citep{sharif2016accessorize,kurakin2016adversarialphysworld}.

In order to counteract adversarial vulnerability, a considerable research effort provided a multitude of defensive approaches for adversarial attacks that can be roughly categorized in two methodologies, that is \emph{rectification} and \emph{adversarial input detection}.
In rectification methods, the goal is to recover the intended output of the model by increasing the robustness of the system, e.g. by trying to remove adversarial perturbation from the input~\citep{li2017adversarial,liao2018defense} or by increasing the robustness of the model itself~\citep{kurakin2016adversarial,papernot2016distillation}.
On the other hand, adversarial detection aims at detecting an occurred attack by analyzing the behavior of the model (without changing it) and signaling anomalous events~\citep{gong2017adversarial,grosse2017statistical,amirian2018trace,metzen2017detecting}. 
Notwithstanding, many of the proposed adversarial detection methods fall prey to strong adversaries too~\citep{carlini2017adversarial}, recent techniques exploiting the training data manifold to ground the predictions of a model~\citep{carrara2019adversarial,papernot2018deep} exhibit good trade-offs between detection performance and resilience to attacks~\citep{sitawarin2019robustness} (as well as tackling a more general problem, that is obtaining good confidence measurements for predictions of deep models~\citep{kendall2017uncertainties}).

While most of the adversarial detection schemes are tested on small or low-resolution benchmarks (such as MNIST and CIFAR datasets), 
this work aims at evaluating one of the aforementioned training-manifold-based adversarial detection methodologies, specifically~\citep{carrara2018adversarial}, in a realistic security-related application that is facial recognition.

Facial recognition systems usually do not usually implement recognition based on deep-learning classifiers but rather follow a similarity-based approach:
deep models are used to extract features from visual facial data, and decisions rely on similarity measurements among those features.
Indeed, standard benchmarks for facial recognition, such as IJB-B~\citep{whitelam2017iarpa} and IJB-C~\citep{maze2018iarpa}, define two evaluation protocols, that is 1:1 Face Verification and 1:N Face Identification.
The former requires to investigate if a person's identity is known or not by comparing its features vector against a database of known identities, while the latter requires to match two images to assess if they belong to the same person or not.

Sticking to those protocols, we provide an analysis of adversarial attacks and further detection in facial recognition systems that implement face identification and verification relying on state-of-the-art deep learning models.
In particular, our contributions are the following:
\textbf{i)} we tested our recently proposed detection technique~\citep{carrara2018adversarial} against classifier attacks, i.e. adversarial samples crafted to fool a state-of-the-art FR neural network acting as a classifier; \textbf{ii)} we generated deep features attacks, using a kNN algorithm as a guidance, to attack a FR system that fulfills the Face Identification task by means of a DL model, acting as a backbone features extractor, followed by a kNN which gives back the query identity based on features similarity; \textbf{iii)} we used deep features attacks to fool a FR system on the Face Verification task, and we showed their superior effectiveness with respect to classifier attacks in fooling such type of system; \textbf{iv)} we used the detectors trained on classifier attacks to detect deep features attacks, thus showing that such approach is generalizable to different types of attacks.\\
The rest of the paper is organized as follows.
In \Cref{related_works}, we briefly reviewed some related works.
In \Cref{adversarial}, we described the algorithms used to craft adversarial examples, while in \Cref{method}, we described the adversarial detection technique used in our study.
In \Cref{experimental_results}, we presented the experimental campaigns that we conducted, and finally, in \Cref{conclusions}, we reported the conclusions of our work.

\section{Related Work} \label{related_works}

\subsection{Adversarial Attacks}
After the seminal work of \cite{szegedy2013intriguing} in which adversarial examples were first studied in DNN, in the last years an exploding growth in studies of adversarial attacks and defenses has been witnessed.
Since the early works, the abundant presence of adversarial examples for standard deep neural networks was confirmed by researchers who proposed multiple crafting algorithms to efficiently find them.
Among the most relevant attacking algorithms available in the literature, there are the box-constrained L-BFGS~\citep{szegedy2013intriguing}, FGSM and its variants~\citep{goodfellow2014explaining,kurakin2016adversarialphysworld,dong2018boosting}, and CW \citep{carlini2017towards}.
We dedicated \Cref{adversarial} for a more detailed review of these algorithms, as we adopted them in this work to generate adversarial examples.

\subsection{Face Recognition Adversarial Attacks}
Face Recognition is among the most important topics in computer  vision. This field has drawn the attention of the scientific community since the early 90s, when \cite{turk1991face}  proposed  the  Eigenfaces  approach. DL models, especially leveraging on the properties of Deep Convolutional Neural Network, started to dominate this field since 2012 reaching  performances  up  to  99.80\%  \citep{wang2018deep}, thus overcoming human performance on this task. Despite the effort in training very robust DL models,  such systems still show some weaknesses.  For example, it has been shown  that  state-of-the-art  face  classifiers  experience  a  performance drop when tested against low resolution images \citep{massoli2019improving}.


Moreover, they are vulnerable to adversarial attacks considering both the black-box~\citep{dong2019efficient} and white-box~\citep{song2018attacks,sharif2016accessorize} settings.\\
Concerning the attacks to face recognition systems, 
\citet{sharif2016accessorize} demonstrated the feasibility and effectiveness of physical attacks by dodging recognition and impersonating other identities using eyeglass frames with a malicious texture.
\citet{dong2019efficient} successfully performed black-box attacks on face recognition models and demonstrated their effectiveness in a real-world deployed system.
Modern attacks on facial recognition systems either exploit generative models obtaining a more natural perturbation~\citep{song2018attacks} or find natural adversarial examples by modifying identity-independent attributes~\citep{qiu2019semanticadv,kakizaki2019adversarial}, such as hair color, makeup, or the presence of glasses.\\
~\citet{pautov2019adversarial} focused on physical world attacks to the LResNet100E-IR FR system. Specifically, they realized adversarial patches that when attached to the area of the face of a person, such as eyes, nose or forehead, or when projected on wearable accessories, allowed the attacker to fool the FR system by leading it to recognize him or her with a different identity.

\subsection{Adversarial Defenses}
Obtaining a system that is robust to adversarial examples turned out to be a challenging and still open task.
The robustness of a model can be increased via adversarial training~\citep{goodfellow2014explaining,huang1511learning} or model distillation~\citep{papernot2015distillation}.
In general, techniques that try to smooth, change, or hide the gradient surface of the model seen by an attacker called gradient-masking defenses, are able to increase the attack effort needed to find an adversarial example, but the enhanced model is still vulnerable to stronger attacks.

Another strategical direction consists of detecting adversarial examples, that is creating robust systems composed by a vulnerable model and a detection system that signals occurring attacks.
Detection subsystems are often implemented as binary detectors that discern authentic and adversarial inputs.
\citet{gong2017adversarial} proposed to train an additional binary classifier that decides whether an input image is pristine or tampered.
\citet{grosse2017statistical} adopted statistical tests in the pixel space to demonstrate the discernibility of adversarial images and proposed to introduce the "adversarial" class in the original classifier which is contextually trained with the model.
Similarly, \citet{metzen2017detecting} proposed a detection subnetwork that relies on intermediate representations constructed by the model at inference time.
However, many detection schemes have been proven to be bypassable~\citep{carlini2017adversarial}.

Novel detection methods rely on the training data manifold for grounding the model prediction and detect anomalies.
\citet{carrara2018adversarial} and \citet{papernot2018deep} showed that a kNN scheme based on intermediate representations of the training set can be used to define a score that measures the confidence of the classification produced by a deep model:
such score can then be used to filter out adversarial examples but also authentic errors occurring.
To cope with the computational cost incurred by a kNN scheme on huge training sets, \citet{carrara2019adversarial} proposed a method that embeds multiple representations in the training space via a distance-based transformation and then performs detection in this space.

%
To our knowledge, the most relevant work that copes with detecting tampered facial recognition is \citet{goswami2019detecting}, in which the authors attacked facial recognition systems in a classification setting and devised a detection approach to decide whether to recover the original input.
In the detection part, they proposed to compare intermediate network activations to their average values defined over a training set, and used layer-wise distances as features in a two-class SVM adversarial detector.
However, their analysis only included the recognition-by-classification setting, while we covered additional real-world settings, such as attacks on kNN identification and verification systems.

\section{Adversarial Attacks}\label{adversarial}
In this section, we described some of the most famous algorithms used for adversarial samples generation.

\subsection{L-BFGS}
~\citet{szegedy2013intriguing} formalised the adversarial attack as an optimization problem that is solved by means of the L-BFGS algorithm.
Specifically, it can be expressed as
\begin{flalign} \label{eq:bfgs}
\mathrm{\underset{r}{min}}\ \ \ \ \ \ \ \ \ & c\ \cdot \parallel r \parallel_2 + \mathcal{L}(x+r,\ t) \nonumber \\
\mathrm{subject\ to} \ \ \ \ &L^m \leq x+r \leq U^m \,,
\end{flalign}
where $[L,\ U]^m$ represents the range of validity for pixel values, and the value of $c>0$ is found by line-search.
The goal of the optimizer is then to find the minimum adversarial perturbation $r$ to the input image $x$ which causes the model to classify $x_{adv}=x+r$ as belonging to the target class $t$.

\subsection{FGSM}
The Fast Sign Gradient Method~\citep{goodfellow2014explaining} (FGSM) is a one-step method in which the optimal max-norm constrained perturbation is found by following the direction of the gradient $\nabla_x J(\theta, x, y)$ of the objective function used to train the DL model with respect to the input image $x \in \mathbb{R}^m$.
The adversarial example is then given by
\begin{flalign} \label{eq:fgsm}
x_{adv} = x + \epsilon \cdot \mathrm{sign}(\nabla_x J(\theta, x, y_{true})) \,,
\end{flalign}
where $\theta$ are the model parameters, $x$ is the input image, $y_{true}$ is its label, and $\epsilon$ is the maximum distortion allowed on the input such that $\parallel x - x_{adv} \parallel_\infty < \epsilon$.

\subsection{BIM}
The Basic Iterative Method~\citep{kurakin2016adversarialphysworld} (BIM) applies the FGSM~\citep{goodfellow2014explaining} attack multiple times with small step size. It is given by
\begin{flalign}
& x^{adv}_0 = x,\ \\ \nonumber 
& x^{adv}_{N+1} = Clip_{x,\epsilon} \big\{x^{adv}_{N} + \alpha \cdot \mathrm{sign}(\nabla_xJ(\theta, x^{adv}_{N}, y_{true}))\big\} \,,
\end{flalign}
where the $Clip(\cdot)$ function clips the values of the pixels at each iteration step to the allowed pixel range, and $\alpha$ is the used step size.


\subsection{MI-FGSM}
The MI-FGSM method~\citep{dong2018boosting} is an iterative procedure that can be generalized to other types of attacks by substituting the current gradient with the accumulated ones from all the previous steps.
The velocity vector in the gradient direction is given by
\begin{flalign} \label{eq:grad}
g_{N+1} = \mu \cdot g_N + 
\frac{J(x^{adv}_N, y)}{\parallel \nabla_x J(x^{adv}_N, y) \parallel_1} \,,
\end{flalign}
where $x^{adv}_0=x$, $g_0=0$, $\mu$ is the decay factor of the running average, and $y$ is the ground truth label.
Subsequently, the adversarial example in the $\epsilon$-vicinity measured by $L_2$ distance is given by
\begin{flalign} \label{eq:l2adv}
x^{adv}_{N+1} = x^{adv}_{N} + \alpha \cdot \frac{g_{N+1}}{\parallel g_{N+1} \parallel_2} \,,
\end{flalign}
where $\alpha=\epsilon/T$ with $T$ being the total number of iterations.

\subsection{Carlini-Wagner Attacks}
\citet{carlini2017towards} (CW) proposed three gradient-based attacks each based on a different distance metric, namely $L_0$,  $L_2$ and  $L_\infty$ attacks.

Given an input $x$ and a target class $t$, different from the original class of the sample, the $L_2$ attack is given by
\begin{flalign} \label{eq:cw2}
&\mathrm{min}\ \left \lVert \frac{1}{2}(\mathrm{tanh}(w)+1) - x \right \rVert^2_2 + 
c \cdot f\left (\frac{1}{2}(\mathrm{tanh}(w)+1) \right ) \nonumber \\
&\mathrm{with} \nonumber \\
&f(x^{adv}) = \mathrm{max}(\mathrm{max} \{Z(x^{adv})_i: i \neq t\} - Z(x^{adv})_t, -k) \,,
\end{flalign}
where $f$ is the objective function, $Z(\cdot)$ are the logits before the softmax layer, $w$ is the variable that represent the adversarial noise in the $\text{tanh}(\cdot)$ space, and $k$ is a parameter that allows to control the confidence with which the misclassification occurs.

Concerning the $L_\infty$ attack, it is not fully differentiable, and the standard gradient descent does not perform well for it.
\autoref{eq:cwinfty} shows the $L_\infty$ version of the attack:
\begin{flalign} \label{eq:cwinfty}
\mathrm{minimize}\ c \cdot f(x + \delta) + \sum_i [(\delta_i - \tau)^+ ] \,.
\end{flalign}

where $\tau$ is a threshold value for the adversarial perturbation. Finally, the $L_0$ attack
is based on the idea of iteratively use $L_2$ to find a minimal set of pixels to be modified to generate an adversarial sample.

\subsection{Deep Features Attack} \label{deep_representation_attack}
All the previous attacks were based on the goal of generating noise which fools the DL model to output a wrong class label for the specific input. \citet{sabour2015adversarial} proposed an approach in which the guiding principle was to create a perturbation of the input image in such a way that its internal representation was similar to the one of a target image.
Starting from a source image $I_s$ and a guide image $I_g$, the goal was to perturb $I_s$ thus generating a new image $I_\alpha$ such that its internal representation, at a layer \emph{k} in the model, $\phi_k(I_\alpha)$, generated by the DL model under attack, had an Euclidean distance from $\phi_k(I_g)$ as small as possible, while $I_\alpha$ remained close to the source $I_s$.
Specifically, $I_\alpha$ was defined to be the solution to the constrained optimization problem
\begin{flalign} \label{eq:deeprepr}
&\emph{I}_\alpha = \mathrm{\underset{\emph{I}}{arg\ min}}
\parallel \phi_k(I) - \phi_k(I_g)\parallel_2^2,
\nonumber \\
&\mathrm{subject\ to} \parallel I - I_s \parallel_\infty < \delta \,,
\end{flalign}
where $\delta$ was the maximum allowed perturbation on each pixel of the source image.

\section{Adversarials Detection Method} \label{method}

\begin{figure}[!h]
\includegraphics[width=\textwidth]{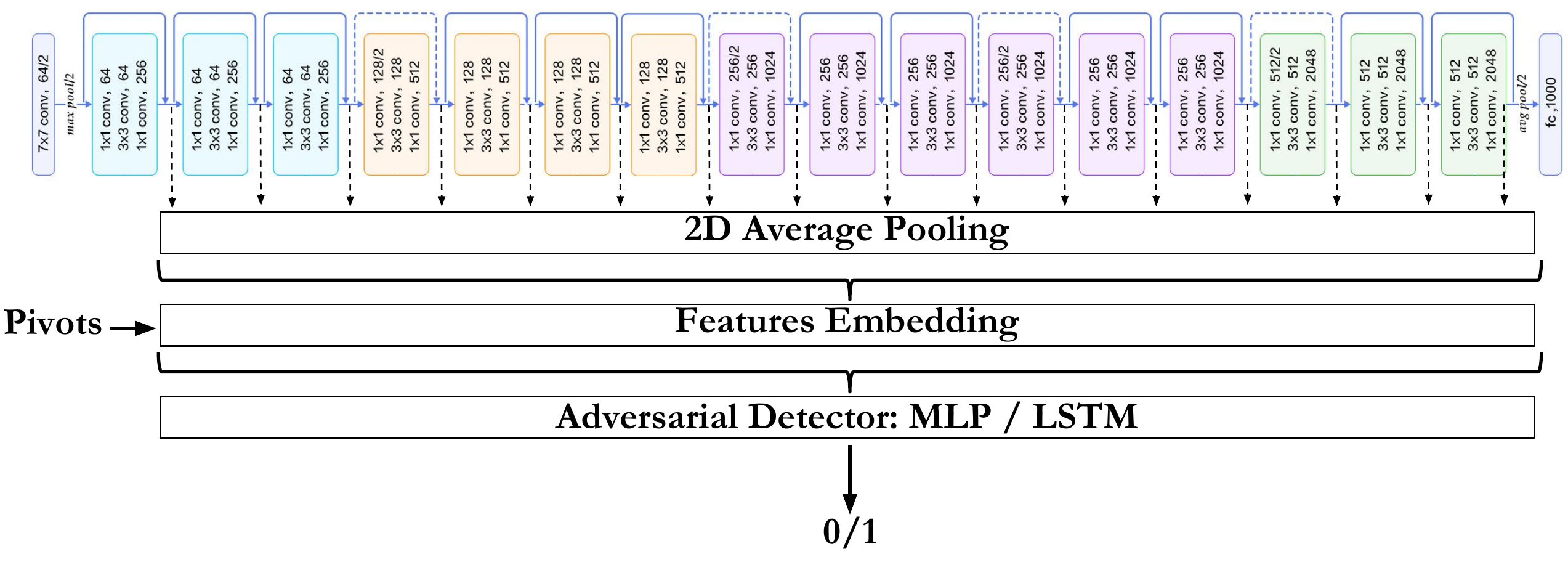}
\caption{Schematic view of the detection algorithm. The deep features are extracted at the end of each of the bottlenecks of the Se-ResNet-50 model.} \label{fig:detector_scheme}
\end{figure}

The ultimate goal of our work was to detect adversarials samples.
In order to accomplish that, we started from the approach we exploited in~\cite{carrara2018adversarial}.
During the forward step of the threatened model, we collected the deep features, at the output of specific layers, to which we subsequently applied an average pooling operation, thus obtaining a single features vector at each selected layer.
Then, we computed the distance among each vector and the class representatives, centroids or medoids, of each class, at each layer, obtaining an embedding which represented the trajectory of the input image in the features space.
Such a trajectory was then fed to a binary classifier that is used as adversarial detector.\\
In our experiments, we used the test set of the VGGFace2~\citep{cao2018vggface2} dataset, which comprises 500 identities, and the state-of-the-art Se-ResNet-50 from~\cite{cao2018vggface2}.
Specifically, we extracted the deep features at the end of each of the 16 bottleneck blocks of the model.
As the adversarial detector, we tested two different architectures: a Multi Layer Perceptron (MLP) and a Long-Short Term Memory (LSTM) network.
The former was made by a hidden layer of 100 units followed by the ReLU non-linear function and a Dropout layer.
The latter had a hidden state size of 100.
In both cases, the output of the detector was fed into a Fully Connected (FC) layer followed by a sigmoid activation function.
A schematic view of the entire system is shown in \autoref{fig:detector_scheme}.

Considering the 16 bottlenecks of the model from which we collected the deep features and the 500 different identities of the dataset, each embedding was then represented by a 8000-dimensional vector.
In this vector, each \emph{i}-th dimension represented the distance between the \emph{i}-th internal representation and a class representative in a specific layer.
\section{Experimental Results} \label{experimental_results}

In this section, we reported the experimental results we have obtained so far.
First, we focused on the detection of the attacks against a state-of-the-art FR model acting as a classifier.
We crafted adversarial inputs by means of known algorithms, and then we trained and tested our detector against them.
Afterwards, we generated deep features (DF) attacks (\Cref{deep_representation_attack}) using a kNN algorithm as guidance for the optimization procedure.

The goal of this approach was to fool a FR system in which the CNN was only used as a features extractor, while the final identity was assigned according to the output of a similarity measurement among deep features.
In our experiments, we considered a system that used a kNN algorithm to assign an identity to the probe image.
This is a typical solution for FR systems to fulfill the Face Identification task. 
Thereafter, we used these deep adversarial features to attack a FR system against the Face Verification task. 
Finally, we used the detectors, trained on the classifier attacks, to detect deep features attacks thus showing the generalization property of the detection approach.

\subsection{Dataset} \label{vggface2_dataset}
As we already stated, in our experiments we employed the test set of the VGGFace2~\cite{cao2018vggface2} dataset.
It comprises 500 identities, with an average of $\sim$340 images for each identity. 

As a first step, for each of the 500 classes of the dataset, we randomly selected 10 images to be used as ``natural" images and 10 to be used for adversarial synthesis.
These selected images were then used to train and test the adversarial detector.

We then split the remaining part of the dataset into train, validation and test sets and used them to train a FC layer on top of the state-of-the-art CNN we used in the study presented in \Cref{micsl_attack}.

\subsection{Classifier Attacks and Detection} \label{micsl_attack}
In the first set of experiments, we focused on the classifier attacks.
As a first step, we replaced the classifier layer of the state-of-the-art facial recognition model~\cite{cao2018vggface2} with a 500-ways FC layer, and we trained it.
To train the model, we used the SGD optimizer with batch size of 256 and a learning rate of $10^{-3}$ halved every time the loss plateaus.
As a preprocessing step, we resized the images so that the shortest side measured 256 pixels.
Afterwards, we randomly cropped a 224x224 region of the image, and we subtracted the average pixel value channel-wise.
For model evaluation, we used the same preprocessing with the exception that the random crop was substituted by a central crop. 

In order to produce adversarial samples, we used the \emph{foolbox}\footnote{\url{https://foolbox.readthedocs.io/en/stable/}} implementation of the BIM~\cite{kurakin2016adversarial}, MI-FGSM~\cite{dong2018boosting}, and CW~\cite{carlini2017towards}, with $L_2$ norm, attacks.
As far as the first two are concerned, we considered a maximum perturbation $\epsilon \in \{0.03, 0.07, 0.1, 0.3\}$,
number of iterations $\in \{30, 50\}$, and for each combination, we considered the targeted and the untargeted versions of the attacks.
The $\epsilon$ values were considered as fractions with respect to the maximum pixel value, which is 255.
Instead, for the CW~\cite{carlini2017towards} attack, we considered the implemented default value of the parameters, i.e. 5 binary search steps and a number of max iterations equals to 1000.
After the adversarial samples generation, we trained the detectors.
The MLP and the LSTM were both trained using the Adam optimizer~\cite{kingma2014adam} for 150 epochs with a batch size of 256 and an initial learning rate ranging from $10^{-4}$ to $10^{-3}$ which was reduced by a factor 10 every time the loss reached a plateau.
Moreover, to balance the sample distribution within mini-batches, we employed a weighted random sampler thus avoiding a bias towards attacks with higher multiplicity.

\begin{figure}[!h]
\includegraphics[width=\linewidth]{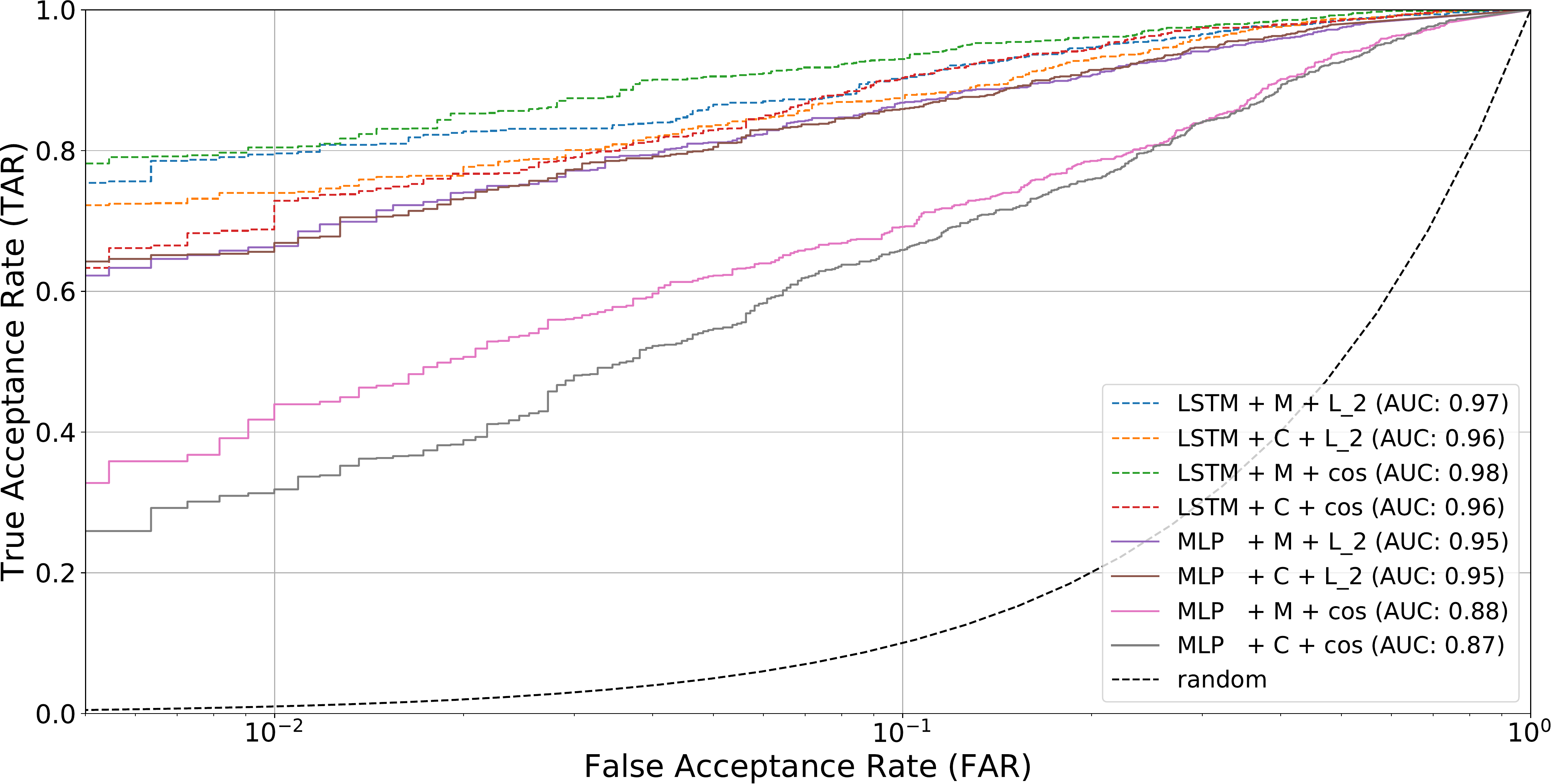} \\
\includegraphics[width=\linewidth]{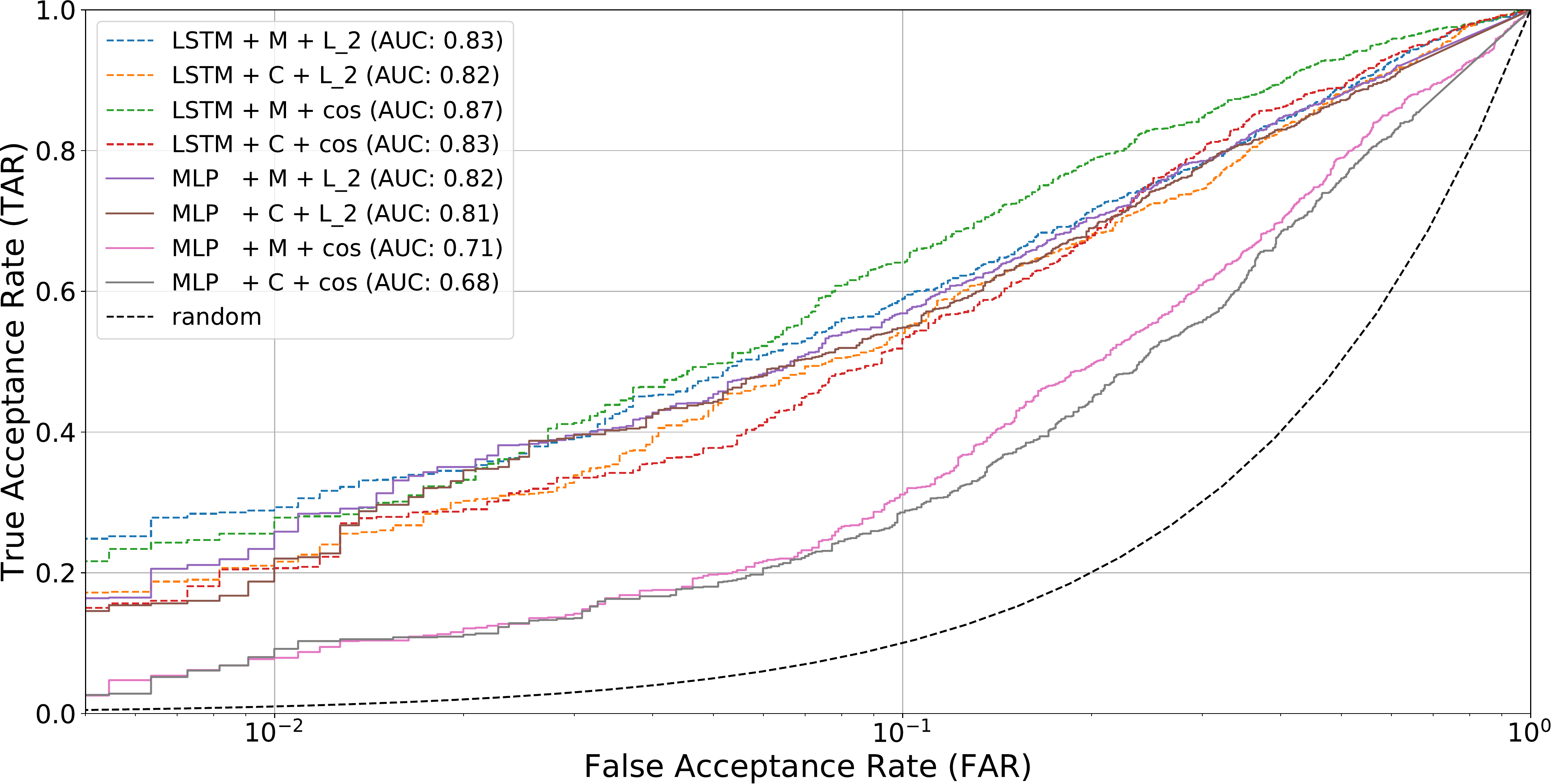}
\caption{ROCs for each model and distance metric combination. Top: targeted attacks. Bottom: untargeted attacks} \label{fig:detector_rocs_aucs}
\end{figure}

In \autoref{fig:detector_rocs_aucs}, we showed the Receiving Operating Characteristics (ROC) curves from the adversarial detection considering targeted and untargeted attacks for each architecture, distance metric, and class representative combination.
As a summary, in \autoref{tab:table_auc_attack_targeted} and \autoref{tab:table_auc_attack_untargeted}, we reported the Area Under the Curve (AUC) values relative to each attack considering their targeted and untargeted versions, respectively.

\begin{table*}[!h]
\caption{Area Under the Curve (AUC) values for each configuration of architecture, pivot-selection and embedding function considering each targeted attack independently. The last column is a summary of the single-attacks AUCs.}
\label{tab:table_auc_attack_targeted}
{\setlength\doublerulesep{0.4pt} 
\begin{tabularx}{\textwidth}{l>{\centering}X>{\centering}X>{\centering}X>{\centering\arraybackslash}X}
\toprule[1pt]\midrule[0.5pt]
Configuration & BIM & CW & MI-FGSM & Macro-AUC \\
\hline
LSTM + M + $L_2$ &  0.977 & 0.871 & 0.986 & 0.944 \\
LSTM + C + $L_2$ &  0.970 & 0.857 & 0.982 & 0.936 \\
LSTM + M + cos   &  \textbf{0.986} & \textbf{0.904} & \textbf{0.991} & \textbf{0.960} \\
LSTM + C + cos   &  0.968 & 0.895 & 0.981 & 0.948 \\
MLP  + M + $L_2$ &  0.964 & 0.793 & 0.979 & 0.912 \\
MLP  + C + $L_2$ &  0.962 & 0.808 & 0.979 & 0.916 \\
MLP  + M + cos   &  0.890 & 0.668 & 0.940 & 0.832 \\
MLP  + C + cos   &  0.868 & 0.720 & 0.915 & 0.834 \\
\midrule[0.5pt]\bottomrule[1pt]
\end{tabularx}
}
\end{table*}

\begin{table*}[!h]
\caption{Area Under the Curve (AUC) values for each configuration of architecture, pivot-selection and embedding function considering each untargeted attack independently. The last column is a summary of the single-attacks AUCs.}
\label{tab:table_auc_attack_untargeted}
{\setlength\doublerulesep{0.4pt} 
\begin{tabularx}{\textwidth}{l>{\centering}X>{\centering}X>{\centering}X>{\centering\arraybackslash}X}
\toprule[1pt]\midrule[0.5pt]
Configuration & BIM & CW & MI-FGSM & Macro-AUC \\
\hline
LSTM + M + $L_2$ & 0.878 & 0.615 & 0.889 & 0.794 \\
LSTM + C + $L_2$ & 0.863 & 0.596 & 0.869 & 0.776 \\
LSTM + M + cos   & \textbf{0.929} & \textbf{0.599}& \textbf{0.930} & \textbf{0.819} \\
LSTM + C + cos   & 0.884 & 0.568 & 0.886 & 0.779 \\
MLP  + M + $L_2$ & 0.885 & 0.559 & 0.882 & 0.775 \\
MLP  + C + $L_2$ & 0.874 & 0.557 & 0.874 & 0.768 \\
MLP  + M + cos   & 0.763 & 0.460 & 0.769 & 0.664 \\
MLP  + C + cos   & 0.730 & 0.467 & 0.739 & 0.645  \\
\midrule[0.5pt]\bottomrule[1pt]
\end{tabularx}
}
\end{table*}

As it was made clear from \autoref{fig:detector_rocs_aucs}, \autoref{tab:table_auc_attack_targeted}, and \autoref{tab:table_auc_attack_untargeted}, the LSTM with the medoids strategy gives the best results.
We can also notice how the CW~\cite{carlini2017towards} algorithm generates samples which are more difficult to be detected with respect to other algorithms. 
Moreover, it is clear how the untargeted attacks are typically more difficult to detect with respect to targeted attacks.
A possible explanation is that these attacks typically find the closest adversarial to the input image, thus an embedding method based on the distance between representation may have difficulties in detecting such attacks.
More details on this intuition were given in \Cref{deep_knn_attacks}.

\begin{figure}[!h]
\includegraphics[width=\textwidth]{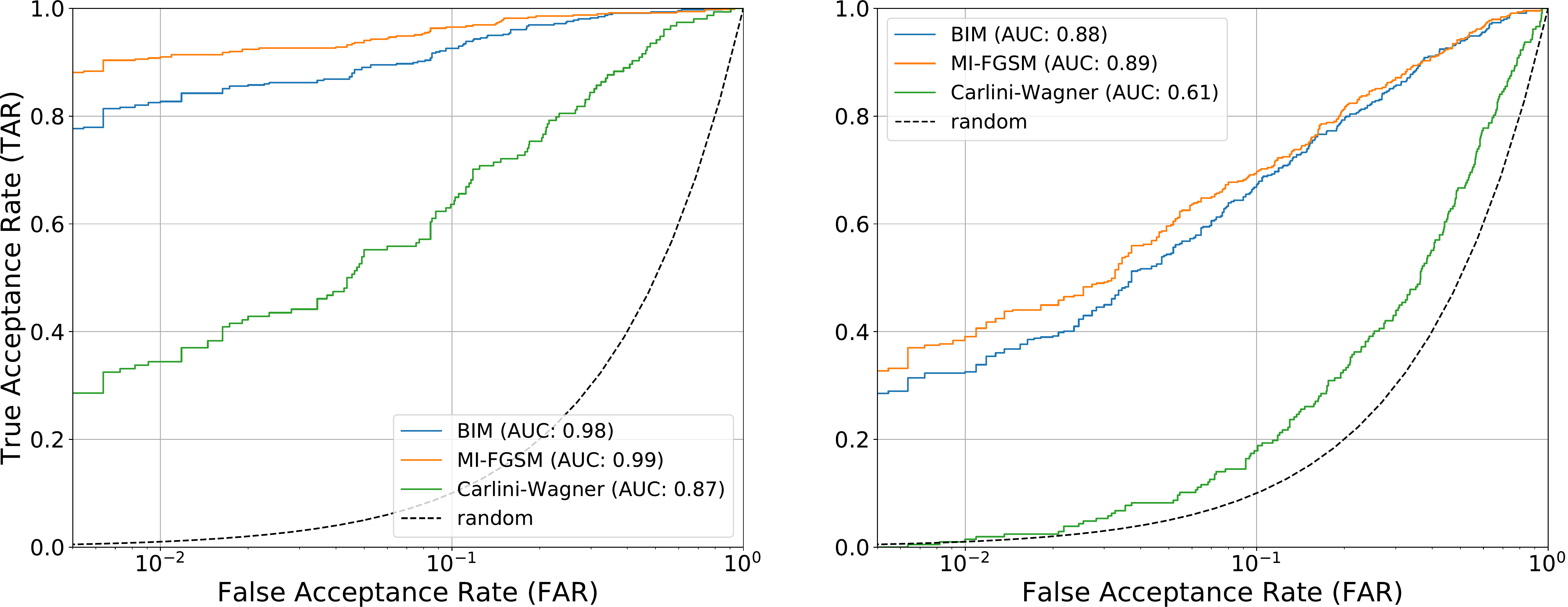}
\caption{ROCs for each attack considering the best trained detector. Left: targeted attacks. Right: untargeted attacks.} \label{fig:best_detector_rocs_aucs}
\end{figure}

Finally, to visually understand the difference among the performances of the best detector on the various attacks, in \autoref{fig:best_detector_rocs_aucs}, we showed the relative ROCs.

According to \autoref{fig:best_detector_rocs_aucs}, it was made even clearer how hard it could be to detect adversarial samples generated by means of the CW~\cite{carlini2017towards} attack, especially considering untargeted attacks.

\subsection{Deep Features Attacks} \label{deep_knn_attacks}
As previously described in \Cref{deep_representation_attack}, it is possible to use the distance among deep representations as a guiding principle to craft adversarial samples instead of wrong label assignment.
Thus, nurturing this idea, we conducted new experiments in which we synthesized adversarial samples by using the distance among deep features as a guidance~\cite{sabour2015adversarial}.

The main idea behind this approach was to emulate a real world application scenario for a FR system in which a learning model is used as a features extractor whose output is used to fulfill the FR task by means of a similarity measurement.

\subsubsection{Face Identification} \label{face_identification}
As a real-world application case, we considered a system relying on a CNN and a kNN algorithm to accomplish the task of Face Identification.
Specifically, what typically happens in this case is that the features vector extracted from the probe face image has to be compared against a database of known identities to identify the person.
Each identity in the database is commonly represented by a template vector, i.e. a vector of features obtained by averaging several deep representations extracted from different images of the same person.
Subsequently, the similarity among the probe vector and the available templates is computed. 
This is what is required, for example, when testing FR model performances on the IJB-B~\cite{whitelam2017iarpa} and IJB-C~\cite{maze2018iarpa} benchmark datasets.
Following this principle, we evaluated the centroids for each of the 500 classes of the dataset.
To conduct our experiments, we used the original state-of-the-art model from~\citet{cao2018vggface2} as a features extractor.

\begin{figure}[!h]
\includegraphics[width=\textwidth]{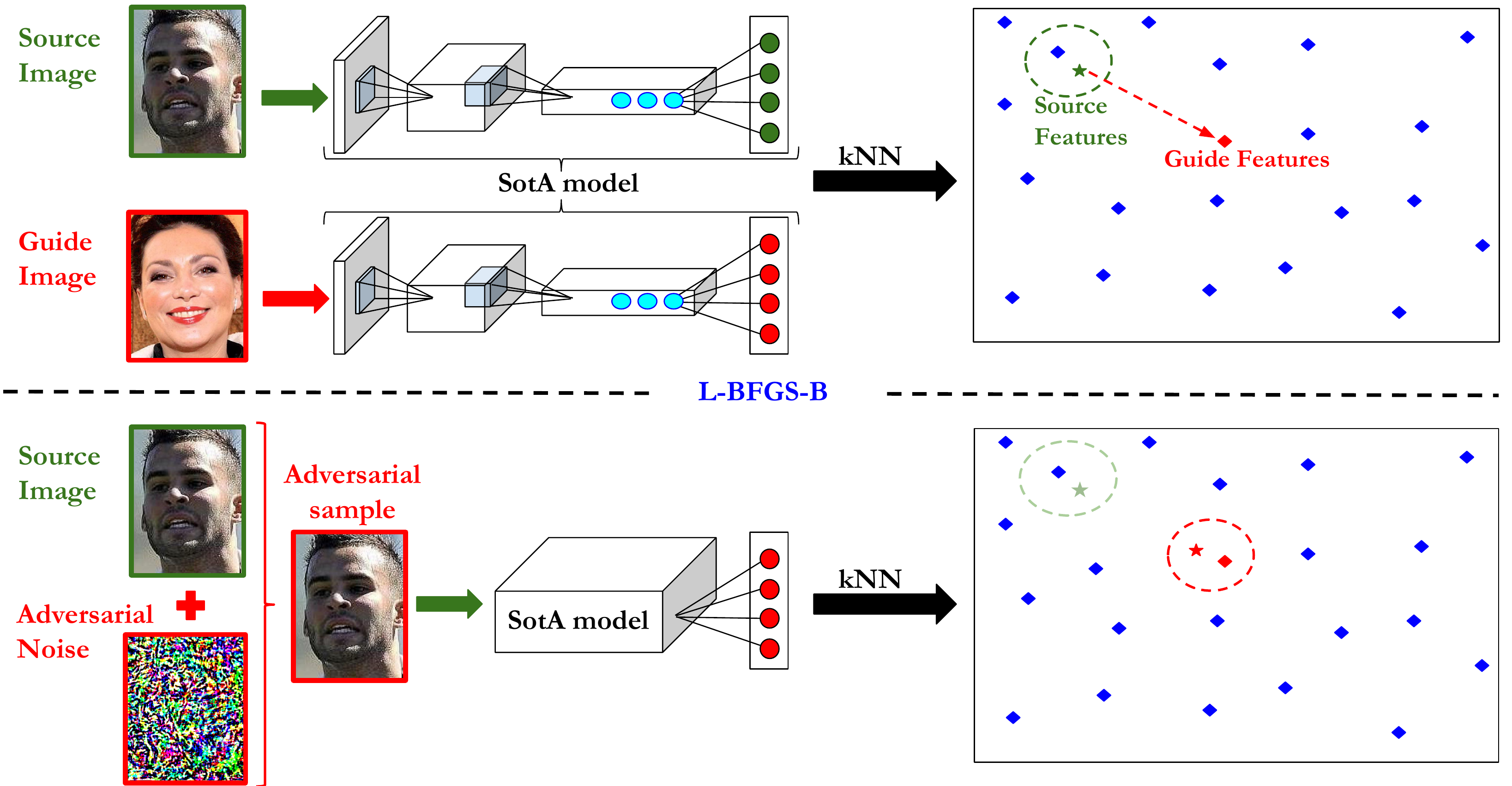}
\caption{Schematic view of the adversarial generation procedure considering a state-of-the-art (SotA) model as features extractor and a kNN to asses the face identity.} \label{fig:deep_knn_sketch}
\end{figure}

\begin{figure}[!h]
\includegraphics[width=\textwidth]{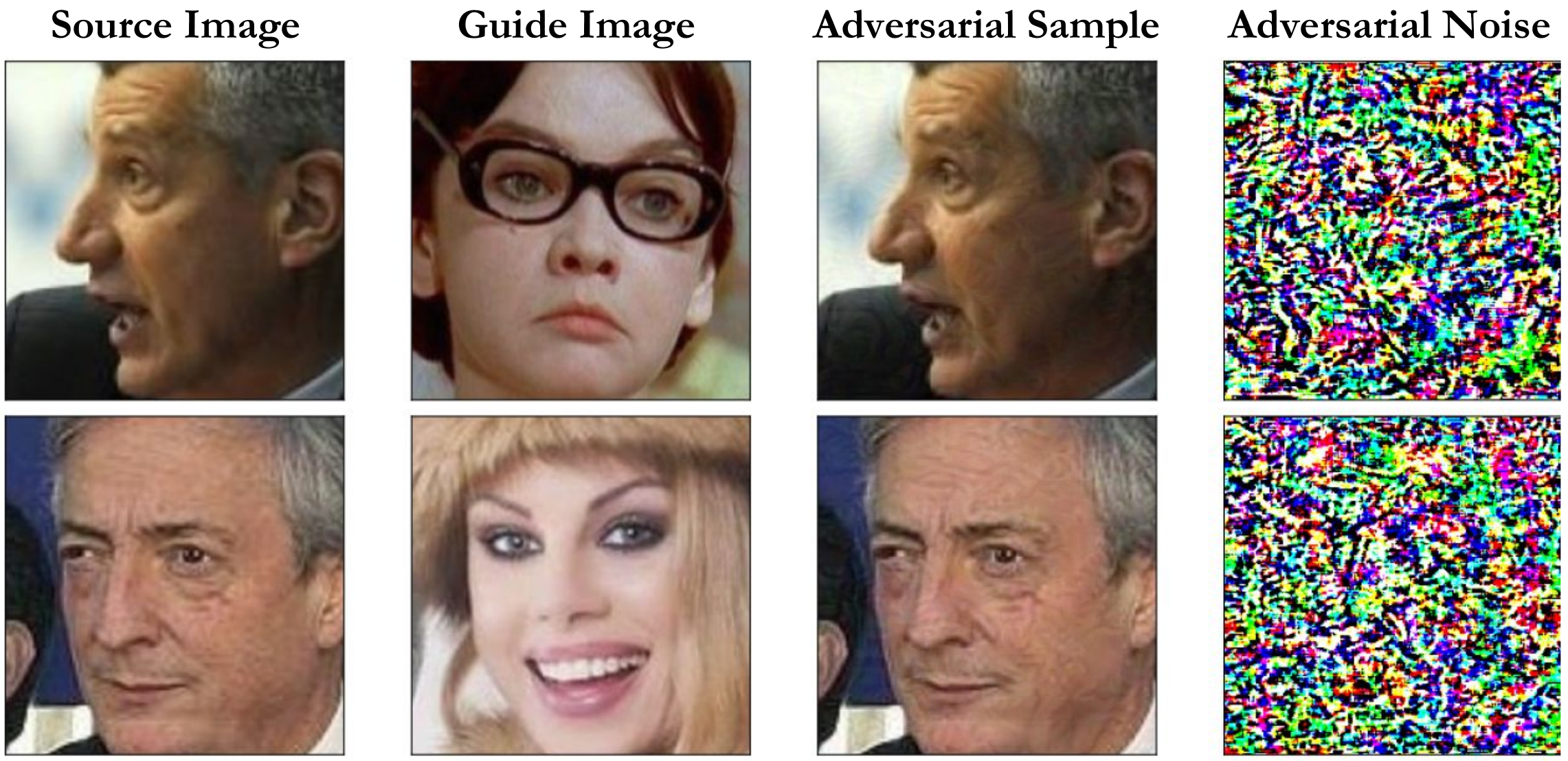}
\caption{Adversarial samples for three different values of the threshold applied while solving the optimization problem. Top: $\delta=5.0$. Middle: $\delta=7.0$. Bottom: $\delta=10.0$} \label{fig:deep_examples}
\end{figure}

The adversarial generation was formulated as an optimization problem~\cite{sabour2015adversarial} solved by using the L-BFGS-B algorithm. Specifically, the constraint was used to set a threshold, $\delta$, on the maximum perturbation on each pixel of the original image as defined in \autoref{eq:deeprepr}.
In order to adapt the adversarial samples generation to our needs, we used a kNN classifier as guidance through the optimization procedure.
The optimization was then stopped once the targeted or untargeted attack's objective were met, that is, the kNN had classified the adversarial as belonging to the guide-image class or it had simply misclassified the face image considering targeted and untargeted attacks respectively. 
A schematic view of the algorithm is shown in \autoref{fig:deep_knn_sketch}.
In our experiments we considered the values of $\delta \in \{5.0, 7.0, 10.0 \}$. An example of adversarial samples generated for each threshold value is shown in \autoref{fig:deep_examples}.
As we can see from \autoref{fig:deep_examples}, the generated images look equal to the original ones, i.e. there is not evident trace of the guide image into the adversarial one.
Considering targeted attacks we obtained a success rate of 95.6\%, 96.2\% and 96.3\% considering a value for $\delta = 5.0, 7.0, 10.0$, respectively. Instead, concerning untargeted attacks we obtained 96.8\% success rate for $\delta = 5.0, 7.0, 10.0$, respectively. In all the samples generations we considered a maximum number of iterations equals to 700.

Differently from what was previously done in \Cref{micsl_attack}, we formulated our approach with the purpose of fooling a FR system which was not based on the simple model classification, but rather on similarities among deep representations. 
Indeed, it is not guaranteed that even if a model misclassifies a face image the adversarial deep representation will be then close enough to the representation of the wrong face class predicted by the model to fool the FR system.

\begin{figure}[!h]
\includegraphics[width=\linewidth]{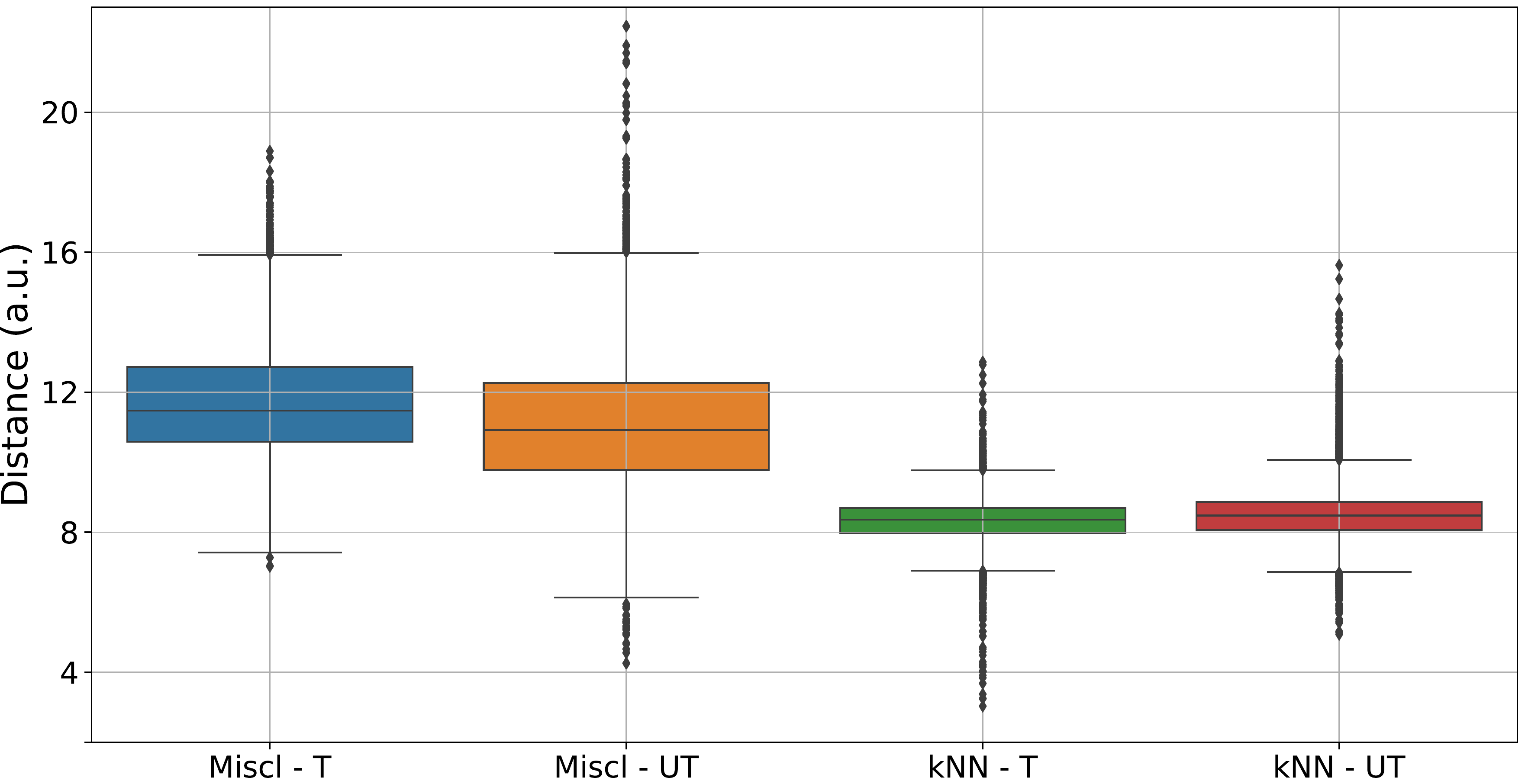}
\caption{Euclidean distance among deep features of adversarial samples and the assigned class centroid for classifier attacks (blue and orange) and kNN-guided attacks (green and red). `` - T" refers to targeted attacks while `` - UT" refers to untargeted attacks.} \label{fig:deep_feat_dist}
\end{figure}

\begin{figure}[!h]
\includegraphics[width=\linewidth]{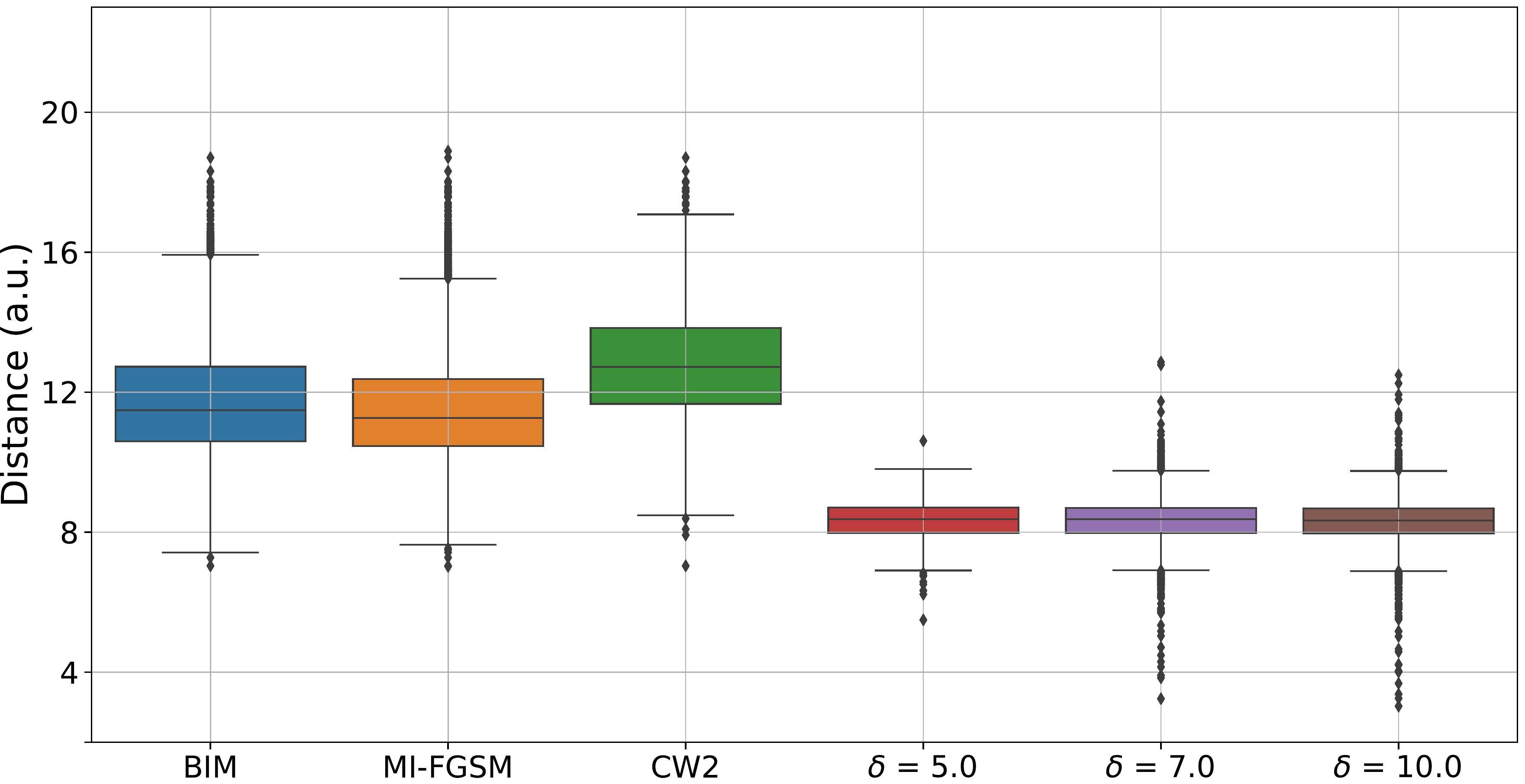}
\caption{Euclidean distance among deep features of adversarial samples and the assigned class centroid considering each targeted attack singularly. The ``$\delta$" values correspond to the maximum $L_\infty$ perturbation allowed, for each pixel, for the kNN-guided attacks.} \label{fig:deep_feat_dist_break_down_t}
\end{figure}

\begin{figure}[!h]
\includegraphics[width=\linewidth]{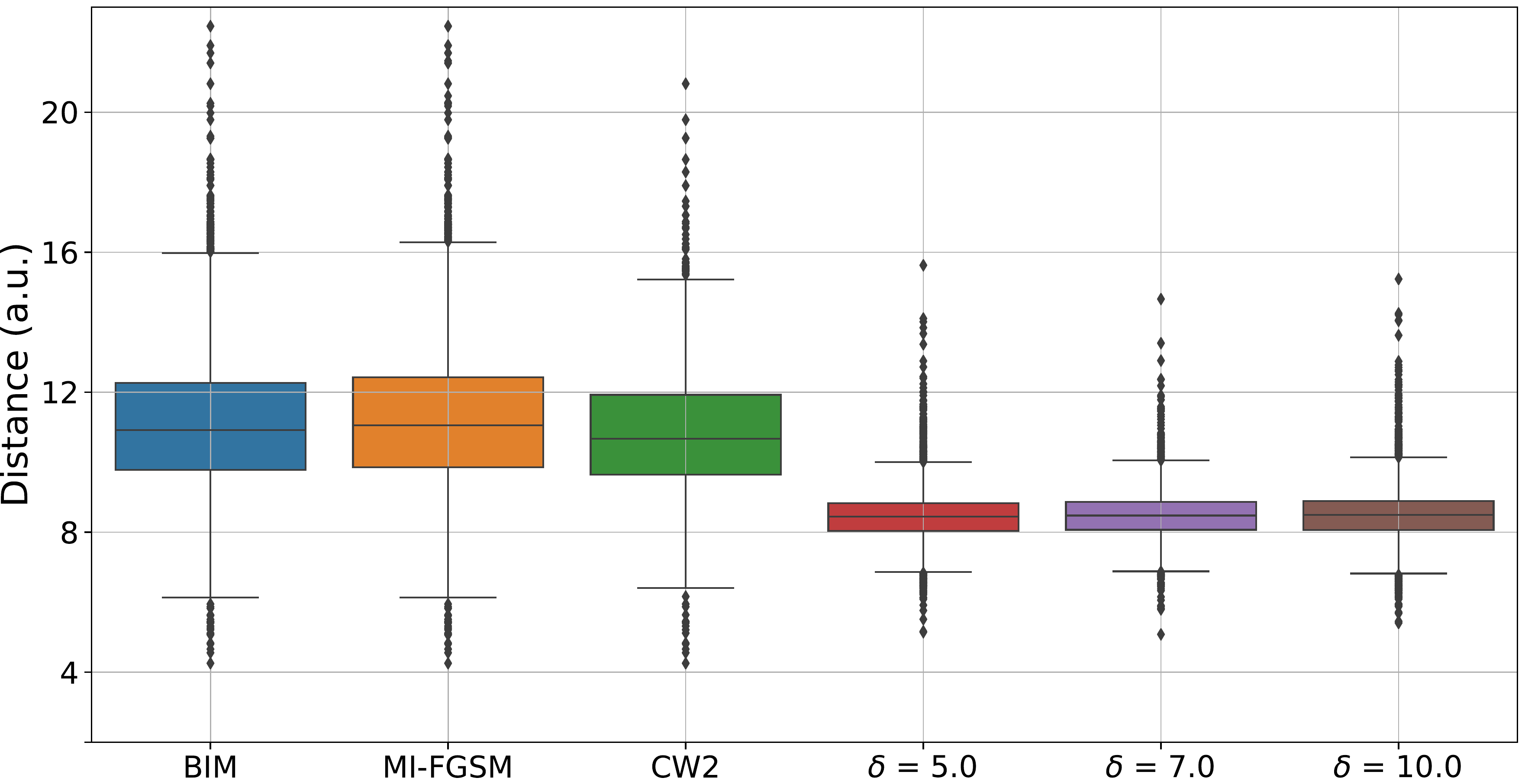}
\caption{Euclidean distance among deep features of adversarial samples and the assigned class centroid considering each untargeted attack singularly. The ``$\delta$" values correspond to the maximum $L_\infty$ perturbation allowed, for each pixel, for the kNN-guided attacks.} \label{fig:deep_feat_dist_break_down_ut}
\end{figure}

In \autoref{fig:deep_feat_dist}, \autoref{fig:deep_feat_dist_break_down_t}, and \autoref{fig:deep_feat_dist_break_down_ut}, we reported some results to justify our intuition.
The figures show the distribution of the distance among the adversarial samples and the centroids of their relative classes considering classifier attacks (BIM~\cite{kurakin2016adversarialphysworld}, MI-FGSM~\cite{dong2018boosting}, and CW~\cite{carlini2017towards}) and DF attacks~\cite{sabour2015adversarial}, using a kNN as guidance. 

As we can see from \autoref{fig:deep_feat_dist}, \autoref{fig:deep_feat_dist_break_down_t}, and \autoref{fig:deep_feat_dist_break_down_ut}, even though classifier attacks are able to fool a CNN model, the distance among them and the centroids of their classes is larger than the one obtained when considering malicious samples generated by means of the deep representation-based attack.
Thus, the latter represents a greater threat, with respect to the former types of attacks, for a FR model.
Moreover, considering targeted and untargeted settings for the classifier attacks, the untargeted attacks are, on average, closer to the class centroids when compared to the targeted ones.
This result supported the observation of the lower detection performance when we tested our detectors against untargeted attacks (\Cref{micsl_attack}).

\begin{figure}[!h]
\includegraphics[width=\linewidth]{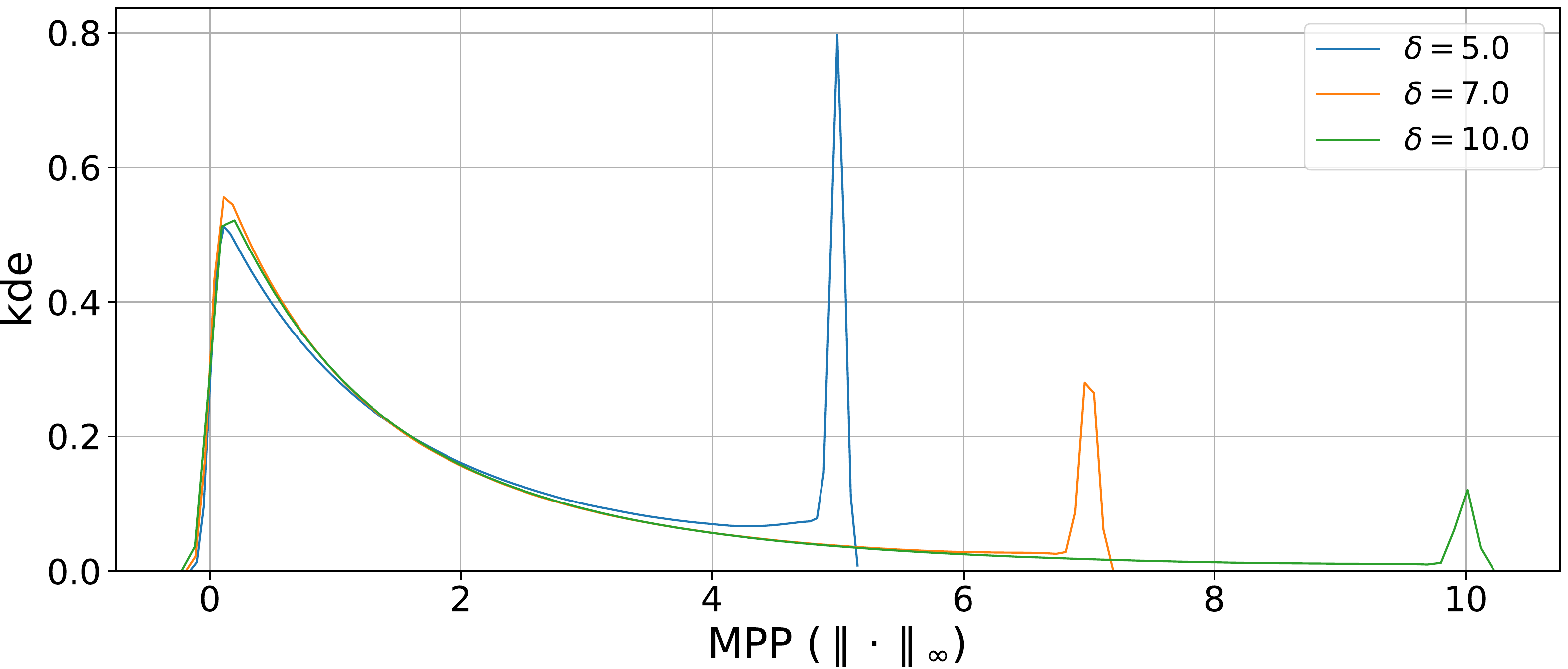}
\caption{Maximum Pixel Perturbation (MPP) distribution considering targeted deep representations with different thresholds} \label{fig:deep_feat_mpp}
\end{figure}

Taking into account \autoref{fig:deep_feat_dist_break_down_t} and \autoref{fig:deep_feat_dist_break_down_ut}, we can also notice that the average distance among the adversarial samples from its class centroid is quite stable among the three different values of the threshold we used when considering malicious images generated with the deep features attack.
This behaviour is supported by the observation that, independently from the threshold applied, the majority of the pixel perturbations are below, in the sense of an $L_\infty$ distance, a threshold of $5.0$.
Thus, a higher threshold has the effect that only a small portion of the image is perturbed above that threshold itself.
Specifically, considering the values of $\delta \in [5.0, 7.0, 10.0]$, the percentage of pixels whose perturbation is within an $L_\infty$ distance of 5.0 is 88.3\%, 85.6\%, 84.7\% respectively.
This behaviour is shown in \autoref{fig:deep_feat_mpp} for targeted attacks.
In the case of an untargeted setting, the results were almost identical.

\subsubsection{Face Verification} \label{face_verification}

In this section, we studied the ability of adversarial attacks to fool a FR system tested against the Face Verification protocol in which two face images are compared to claim if they belong to the same identity or not.
In the DL context, such decision is typically based upon similarity measurements among deep features extracted from the input faces. Specifically, once the model has been trained, the ROC curve is typically evaluated and a threshold value is chosen to be used as a reference value. Then, two faces are said to belong to the same identity if their similarity score exceeds the predefined threshold.
An example of a real world application scenario of this kind of (such a) protocol is a restricted access area control system.
Since in this type of applications the False Positives pose a greater threat than the False Negatives, it is important to evaluate the ROC curve down to very low values of the False Acceptance Rate (FAR). Such a demand translates into the requirement of evaluating the similarity scores among a larger number of negative pairs with respect to the positive ones.

Even in this case, the architecture of a FR system was made by a features extractor and a module which worked out the similarity measurements.
As a features extractor we used the state-of-the-art model from \citet{cao2018vggface2}, while we considered the cosine among features vectors as similarity measurement.

After training the model, we obtained a ROC curve with an AUC value equals to 99.03\%. Then, we used the Equal Error Rate (EER) threshold which we found equal to 0.448 as a threshold value for the similarity measurement.
At this point, we hypothesized two possible scenarios for the adversarial attacks:

\begin{itemize}
    \item \emph{Impersonation Attack}. In this case, we wanted to fool the system by leading it to falsely predict that two face images belonged to the same identity.
    This situation emulated the case in which an intruder intends to enter a restricted area or, in a more general case, when someone is made recognizable as a different person.
    \item \emph{Evading Attack}. This case is the opposite of the previous one, i.e., we wanted the system not to recognize a person by saying that the two images belonged to different identities.
     This circumstance imitated the condition of someone whose identity is made unrecognizable.
\end{itemize}
From the FR system perspective, in the former we needed the two images, which belonged to different people, to be ``equal enough", i.e., their similarity measurement had to be above the threshold we had previously defined, while in the latter, we needed the two images to be ``distant enough", i.e. below the threshold.

\begin{figure*}[!h]
\includegraphics[width=\textwidth]{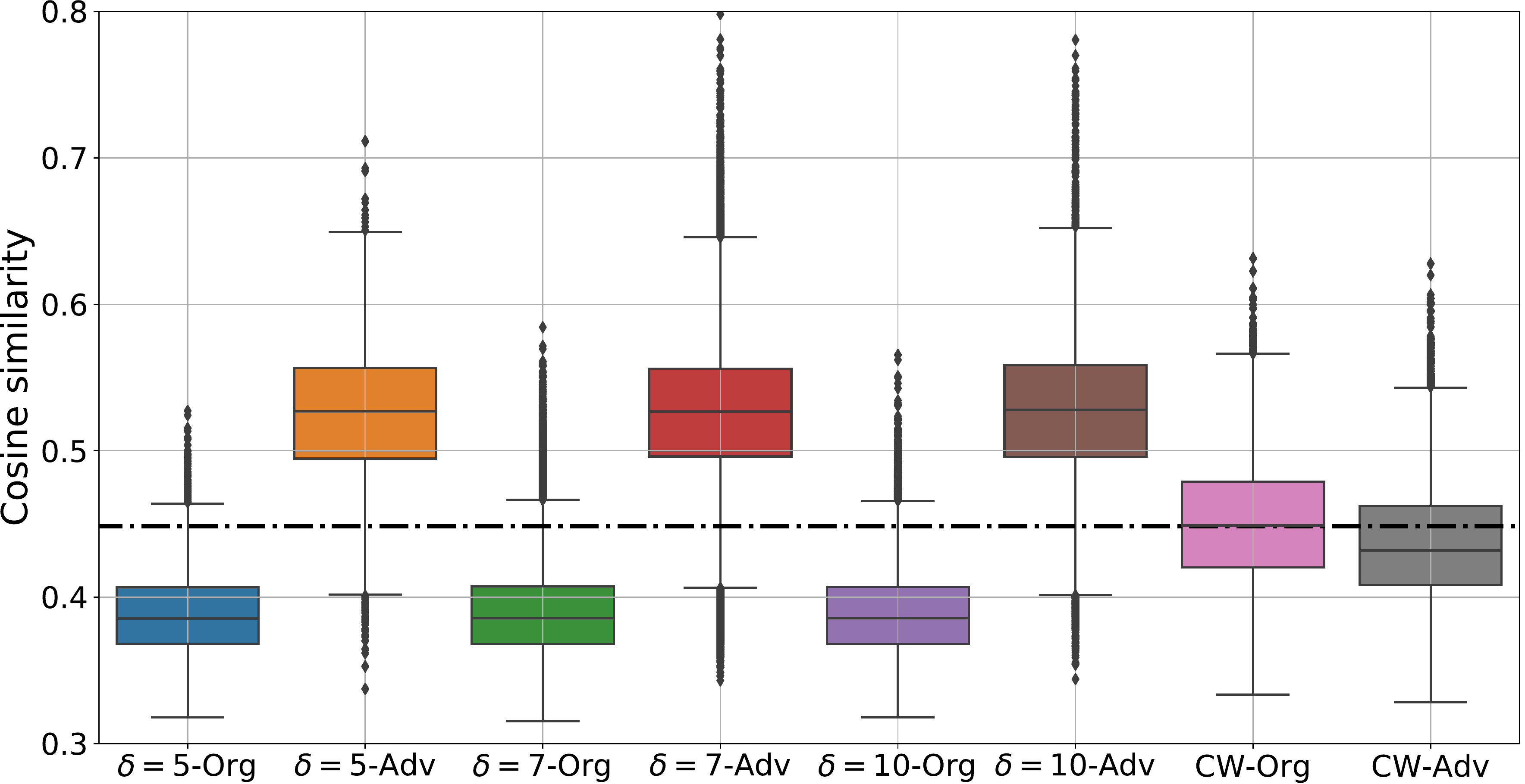} \\
\includegraphics[width=\textwidth]{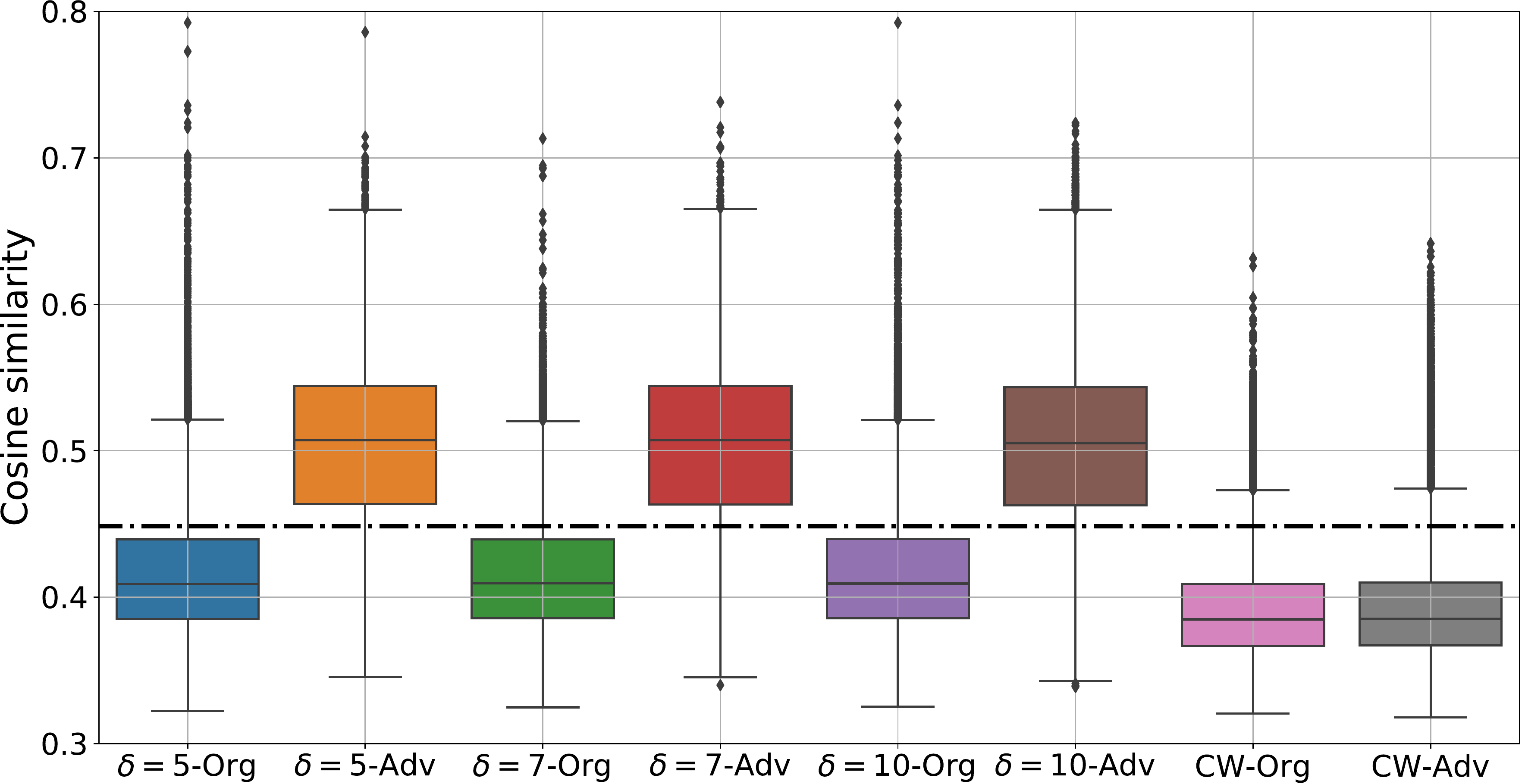}
\caption{Cosine similarity distribution for kNN-guided and CW attacks in the \emph{Impersonation Attack} scenario. Top: targeted attacks. Bottom: untargeted attacks. ``- Org" refers to the cosine among natural images while ``- Pred" refers to the cosine between the natural image and the adversarial one. The dash-pointed line represents the EER threshold.} \label{fig:deep_adv_pred_same_targeted_untargeted}
\end{figure*}

To carry out these experiments, we consider the CW \cite{carlini2017towards} attacks and the kNN guided ones in the targeted and untargeted settings.
What we expected was the deep representation attacks to be more effective with respect to the classifier attacks.
The results for the \emph{Impersonation Attack} scenario are reported in \Cref{fig:deep_adv_pred_same_targeted_untargeted}.
In this case, we considered what follows:
first, we randomly selected negative matches, and we kept the second image fix, i.e. we analysed pairs of faces $(x,\ x^{-})$ where $x^-$ was an image from a different class of $x$.
Then, we looked upon an adversarial image, whose adversarial class corresponded to the one of $x^-$, and used it in place of the first image of the match, i.e. we accounted the pairs $(x_{adv},\ x^{-})$ where $x_{adv}$ was an adversarial sample, crafted from $x$, whose adversarial class was the same as the $x^{-}$ one.
Then, we considered two similarity measurements: 
\begin{itemize}
    \item ``Original" which represents the value of the cosine among the deep features of $x$ and $x^-$;
    \item ``Adversarial" which represents the cosine between the deep features of $x_{adv}$ and $x^-$.
\end{itemize}

In \Cref{tab:pred_same_targeted_untargeted}, we reported the percentage of matches which overcame the EER threshold, before and after the attacks, considering the targeted and untargeted settings.

\begin{table}[!h]
\caption{Percentage of matches which overcame the EER threshold, before and after the attacks, considering the targeted and untargeted settings.}
\label{tab:pred_same_targeted_untargeted}
{\setlength\doublerulesep{0.4pt} 
\begin{tabularx}{\textwidth}{l>{\centering}X>{\centering}X>{\centering}X>{\centering\arraybackslash}X}
\toprule[1pt]\midrule[0.5pt]
\multicolumn{1}{c}{} & \multicolumn{2}{c}{\textbf{Targeted}} & \multicolumn{2}{c}{\textbf{Untargeted}} \\ 
 & Original & Adversarial & Original & Adversarial  \\ \cline{2-5} 
$\delta=5$ & 4.0 & 92.8 & 19.9 & 81.5 \\ 
$\delta=7$ & 4.0 & 93.9 & 19.9 & 81.6 \\ 
$\delta=10$ & 4.4 & 93.6 & 19.8 & 81.1 \\ 
CW \cite{carlini2017towards} & 50.5 & 34.9 & 8.0 & 9.3 \\ 
\midrule[0.5pt]\bottomrule[1pt]
\end{tabularx}
}
\end{table}

As we can observe from \Cref{tab:pred_same_targeted_untargeted}, the DF attacks~\cite{sabour2015adversarial} (kNN-guided) are much more effective in pushing the similarity between the adversarials and the natural images above the recognition threshold.
Such conclusion holds for targeted and untargeted attacks.
Instead, the behaviour of the CW \cite{carlini2017towards} was unpredictable in this set up, thus we can conclude that even though the CW \cite{carlini2017towards} algorithm is among the strongest ones concerning classification attacks, although it is not very effective against the Face Verification protocol. 

\begin{figure*}[!h]
\includegraphics[width=\textwidth]{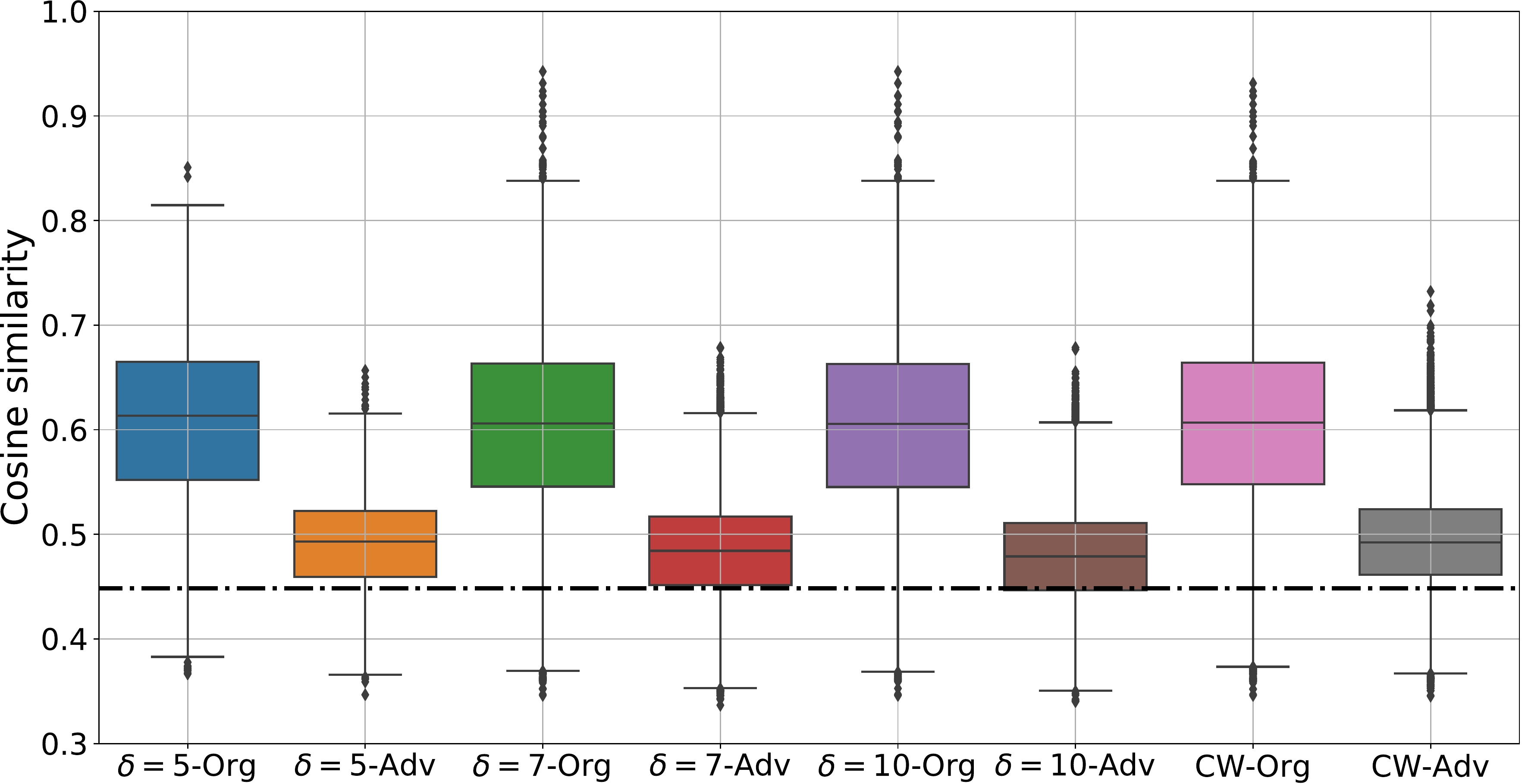} \\
\includegraphics[width=\textwidth]{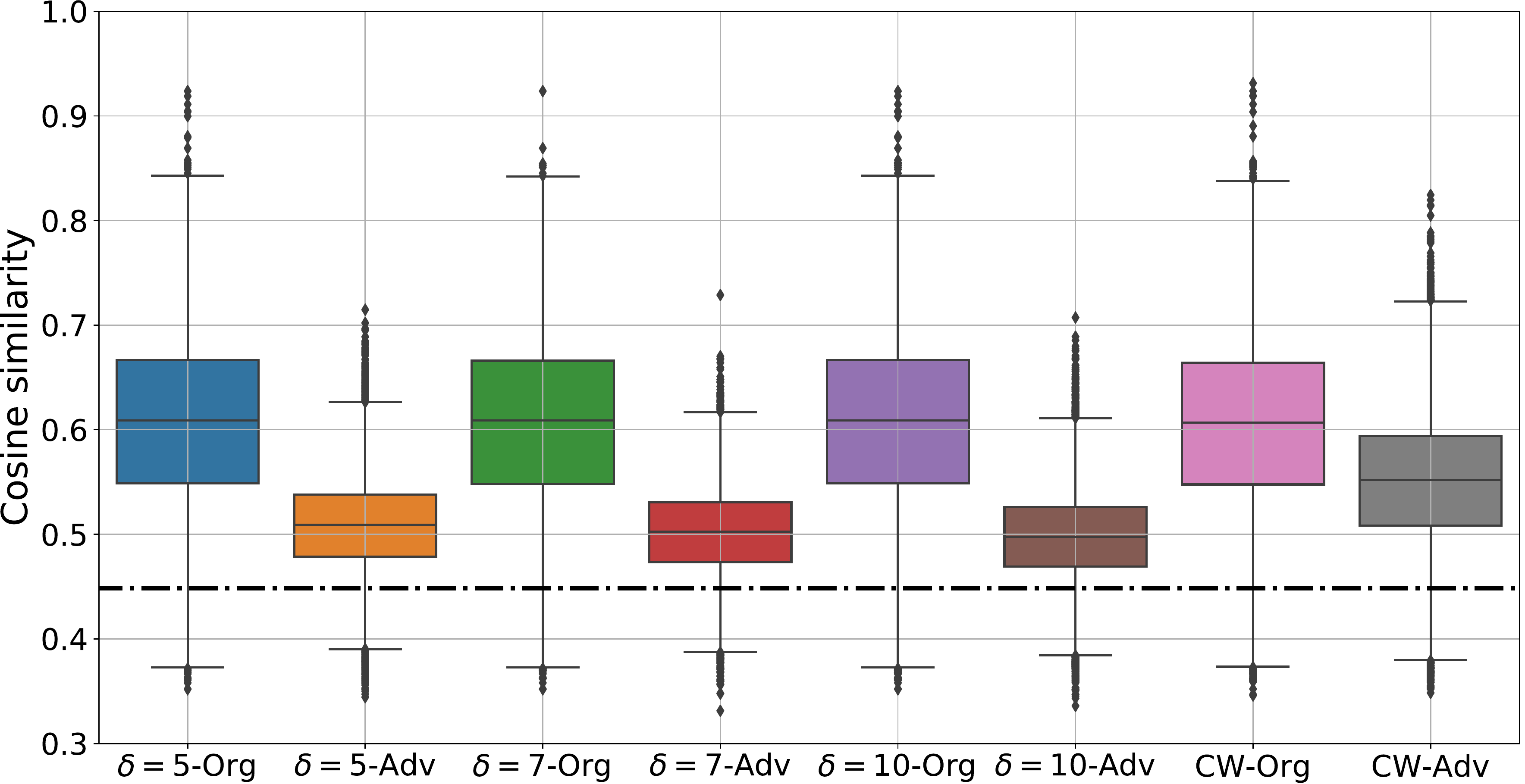}
\caption{Cosine similarity distribution for kNN-guided and CW attacks in the \emph{Evading Attack} scenario. Top: targeted attacks. Bottom: untargeted attacks. ``- Org" refers to the cosine among natural images while ``- Pred" refers to the cosine among the natural image and the adversarial one. The dash-pointed line represents the EER threshold.} \label{fig:deep_adv_pred_not_same_targeted_untargeted}
\end{figure*}

As far as the \emph{Evading Attack} scenario is concerned, the results are reported in \Cref{fig:deep_adv_pred_not_same_targeted_untargeted}.
Differently from the previous case, we started by collecting positive matches, i.e. pairs of images $(x,\ x^+)$ in which $x$ and $x^+$ belonged to the same class, and then we substituted $x$ with one of its adversarial, $x_{adv}$, whose class was different from the $x^+$ one.
Thus, we obtained the following similarity measurements:
 \begin{itemize}
    \item ``Original" which represents the value of the cosine among the deep features of $x$ and $x^+$;
    \item ``Adversarial" which represents the cosine between the deep features of $x_{adv}$ and $x^+$.
\end{itemize}
As we explained before, in this scenario the purpose of the attack was to push the similarity below the operational level of the FR system. 

In \Cref{tab:pred_not_same_targeted_untargeted}, we reported the percentage of the matches that were below the EER threshold, before and after the attacks, considering the targeted and untargeted settings.

\begin{table}[!h]
\caption{Percentage of matches which are below the EER threshold, before and after the attacks, considering the targeted and untargeted settings.}
\label{tab:pred_not_same_targeted_untargeted}
{\setlength\doublerulesep{0.4pt}
\begin{tabularx}{\textwidth}{l>{\centering}X>{\centering}X>{\centering}X>{\centering\arraybackslash}X}
\toprule[1pt]\midrule[0.5pt]
\multicolumn{1}{c}{} & \multicolumn{2}{c}{\textbf{Targeted}} & \multicolumn{2}{c}{\textbf{Untargeted}} \\ 
& Original & Adversarial & Original & Adversarial  \\ \cline{2-5}  
$\delta=5$ & 5.4 & 18.9 & 4.1 & 9.9 \\ 
$\delta=7$ & 4.3 & 22.9 & 4.3 & 11.4 \\ 
$\delta=10$ & 4.3 & 26.2 & 4.1 & 12.5 \\ 
CW \cite{carlini2017towards} & 4.0 & 17.1 & 4.0 & 6.2 \\ 
\midrule[0.5pt]\bottomrule[1pt]
\end{tabularx}
}
\end{table}

By observing \Cref{tab:pred_not_same_targeted_untargeted}, it was clear that the DF attacks~\cite{sabour2015adversarial} (kNN-guided) were more effective that the CW \cite{carlini2017towards} attacks in this case too.
We can noticed that on average the targeted attacks performed better than the untargeted ones, which was an expected behaviour since an untargeted attack ended as soon as the the adversarial is associated with a different identity, therefore it would not have gone any further from the original image.

\subsubsection{Detection}

\begin{figure}[!h]
\includegraphics[width=\linewidth]{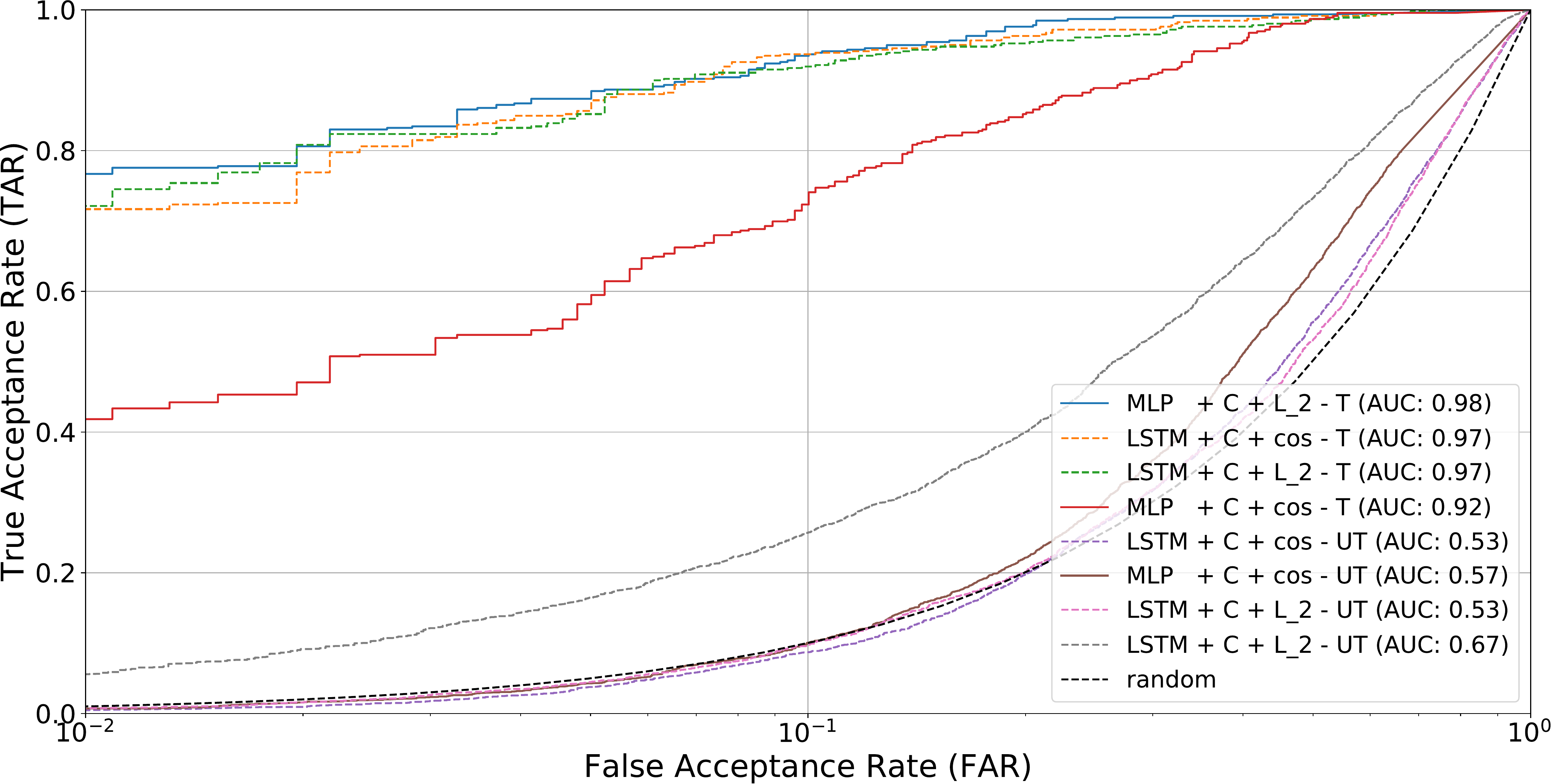} \\
\includegraphics[width=\linewidth]{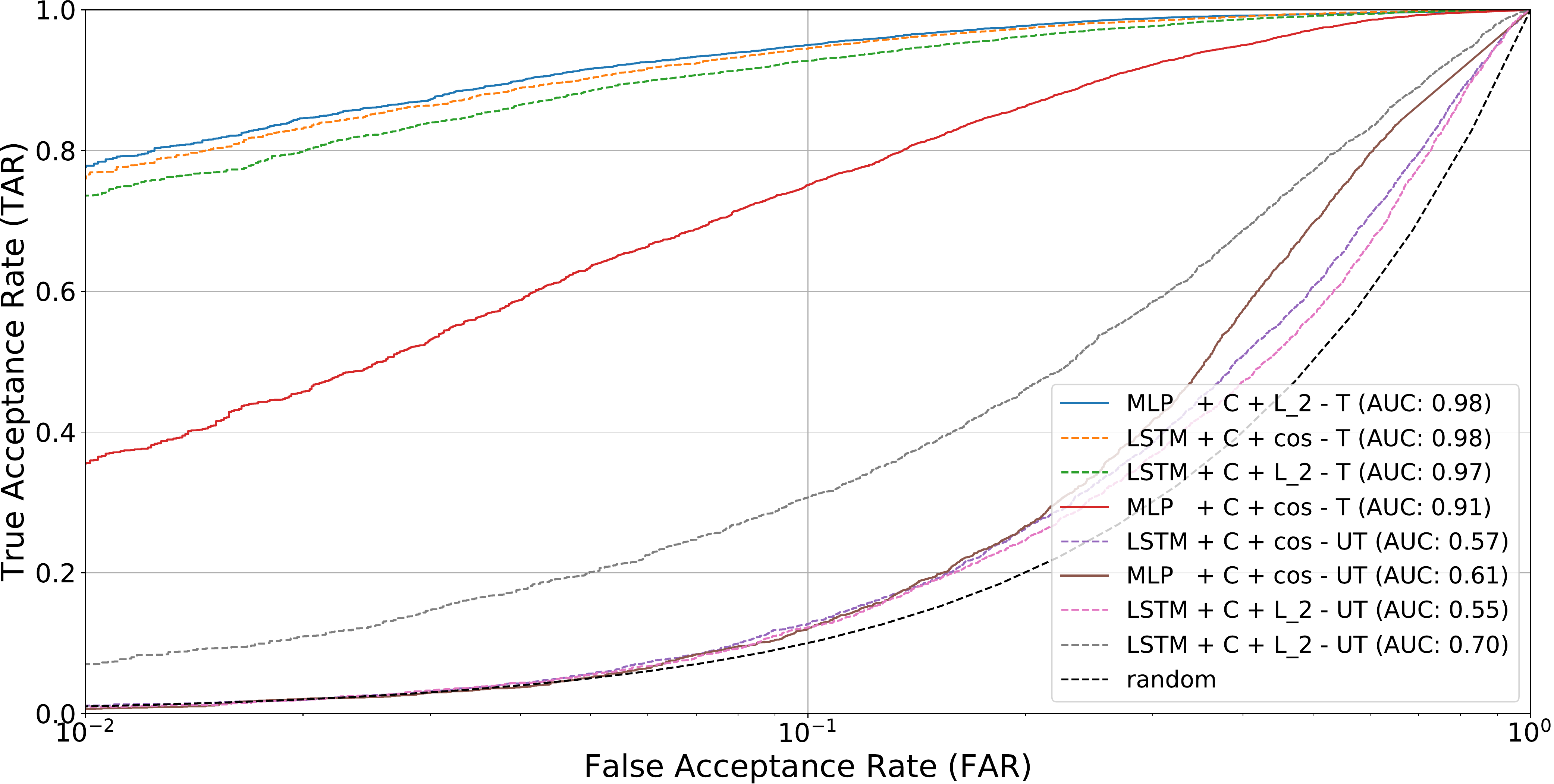}
\caption{ROCs for the best models considering adversarial attacks generated with $\delta=5.0$ (top) and $\delta=10.0$ (bottom). `` - T" refers to targeted attacks while `` - UT" refers to untargeted attacks.} \label{fig:rocs_deep_repr_delta_5}
\end{figure}

Finally, we tested our detectors, trained on classifier attacks (\Cref{micsl_attack}), on the newly generated adversarial samples.
The resulting ROC curves, according to a threshold of $\delta=5$ and $\delta=10$ for targeted and untargeted attacks configurations, are shown in \autoref{fig:rocs_deep_repr_delta_5}.
We did not report the ROC for the case $\delta=7$ since the results were almost identical to the case with $\delta=10$.

As a summary, the AUC values were reported in \autoref{tab:tsb_deep_repr_t} and \autoref{tab:tsb_deep_repr_ut} for targeted and untargeted attacks, respectively.
\begin{table}[!h]
\caption{AUC values for the best performing detectors for each threshold value considered in our experiments in the case of targeted attacks.}
\label{tab:tsb_deep_repr_t}
{\setlength\doublerulesep{0.4pt} 
\begin{tabularx}{\linewidth}{l>{\centering}X>{\centering\arraybackslash}X}
\toprule[1pt]\midrule[0.5pt]
Configuration & AUC & $\delta$ \\
\hline
MLP  + $L_2$ & \textbf{0.976} & 5 \\
LSTM + cos   & 0.972 & 5 \\
LSTM + $L_2$ & 0.969 & 5 \\
MLP  + cos   & 0.915 & 5 \\ \hline
MLP  + $L_2$ & \textbf{0.977} & 7 \\
LSTM + cos   & 0.975 & 7 \\
LSTM + $L_2$ & 0.968 & 7 \\
MLP  + cos   & 0.908 & 7 \\ \hline
MLP  + $L_2$ & \textbf{0.980} & 10 \\ 
LSTM + cos   & 0.978 & 10 \\
LSTM + $L_2$ & 0.972 & 10 \\
MLP  + cos   & 0.915 & 10 \\
\midrule[0.5pt]\bottomrule[1pt]
\end{tabularx}
}
\end{table}
\begin{table}[!h]
\caption{AUC values for the best performing detectors for each threshold value considered in our experiments in the case of untargeted attacks.}
\label{tab:tsb_deep_repr_ut}
{\setlength\doublerulesep{0.4pt} 
\begin{tabularx}{\linewidth}{l>{\centering}X>{\centering\arraybackslash}X}
\toprule[1pt]\midrule[0.5pt]
Configuration & AUC & $\delta$ \\
\hline
LSTM + cos   & 0.530 & 5 \\
MLP  + $L_2$ & 0.492 & 5 \\
MLP  + cos   & 0.571 & 5 \\
LSTM + $L_2$ & \textbf{0.671} & 5 \\ \hline
LSTM + cos   & 0.541 & 7 \\
MLP + $L_2$ & 0.481 & 7 \\ 
MLP  + cos   & 0.596 & 7 \\
LSTM + $L_2$ & \textbf{0.688} & 7 \\ \hline
LSTM + cos   & 0.573 & 10 \\
MLP + $L_2$ & 0.467 & 10 \\
MLP  + cos   & 0.609 & 10 \\
LSTM + $L_2$ & \textbf{0.700} & 10 \\
\midrule[0.5pt]\bottomrule[1pt]
\end{tabularx}
}
\end{table}
According to the results shown in \autoref{fig:rocs_deep_repr_delta_5}, \autoref{tab:tsb_deep_repr_t}, and \autoref{tab:tsb_deep_repr_ut}, we could see that, even though the adversarial detectors were trained on different attacks, they displayed high performances in detecting kNN-guided attacks too.
This result has a relevant significance since it means that, despite the different attacks' objectives, adversarial samples share some common behaviours in the inner layers of a deep model.
Moreover, it highlights the generalization capacity of our detection approach. 
Furthermore, we can acknowledged that while the AUC values were very close for the targeted attacks, in the case of untargeted attacks the LSTM performed considerably better than the MLP. 

\section{Conclusions}\label{conclusions}
Adversarial samples represent a serious threat to DL models, epsecially as they set a serious limitation especially on the use of learning models in sensitive applications.
Despite the scientific community's effort in trying to train robust NN, a knowledgeable attacker usually succeeds in finding ways to attack a model. 

Except for the adversarial training, another approach to enhance the robustness of AI-based systems to the adversarial threat is detecting these malicious inputs. In several previous studies the properties of the offensive samples are exploited in order to detect them. Compared to adversarially training a model, the detection of these images has several advantages, e.g. it does not require to re-train any model nor it does not require to specifically design new training strategies to flatten the model loss manifold.

In light of these facts, we proposed our study on the detection of the adversarial samples.
Specifically, we exploited the different behaviour of adversarial samples in the inner layers of a DL model with respect to natural images.

We conducted our experiments in the context of Face Recognition, for which we crafted adversarial samples considering wrong-label assignment and deep representation distance as objectives in the targeted and untargeted settings.
We first considered the NN acting as a classifier, and then we conducted our attacks against a FR system in which the learning model was employed as features extractor. 
As far as the classifier attacks are concerned, the best detector reached an AUC value of 99\% on the adversarial detection task.

The results obtained from the deep features attacks against a FR system are even more interesting.
In this case, we considered a more realistic application scenario for a FR system in which the DL model was used as a features extractor, and the final task was accomplished by means of similarity measurements among the descriptors vectors.
Specifically, we observed that
i) classifier attacks are much less effective in fooling a FR system;
ii) the detectors, trained on the first type of attacks, reached an AUC value of 98\% and 70\% for the deep representation attacks, which they had never seen before, for targeted and untargeted attacks, respectively.
These last results are of great impact considering the idea of an ``universal" adversarial detector.
Moreover, this also means that, despite the different objectives of the various kind of attacks, they actually share some common properties that can, or perhaps should, be exploited to recognize adversarial attacks and build more robust systems without the need to periodically change the model to increase its robustness.

\section*{Acknowledgments} This work was partially supported by the AI4EU project, funded by the EC (H2020 - Contract n. 825619), and by Automatic Data and documents Analysis to enhance human-based processes (ADA) project, CUP CIPE D55F17000290009.
We gratefully acknowledge the support of NVIDIA Corporation with the donation of the Titan V GPU used for this research.



\begin{thebibliography}{46}
\providecommand{\natexlab}[1]{#1}
\providecommand{\url}[1]{\texttt{#1}}
\providecommand{\urlprefix}{URL }
\expandafter\ifx\csname urlstyle\endcsname\relax
  \providecommand{\doi}[1]{doi:\discretionary{}{}{}#1}\else
  \providecommand{\doi}[1]{doi:\discretionary{}{}{}\begingroup
  \urlstyle{rm}\url{#1}\endgroup}\fi
\providecommand{\bibinfo}[2]{#2}

\bibitem[{Krizhevsky et~al.(2012)Krizhevsky, Sutskever, and
  Hinton}]{NIPS2012_4824}
\bibinfo{author}{A.~Krizhevsky}, \bibinfo{author}{I.~Sutskever},
  \bibinfo{author}{G.~E. Hinton}, \bibinfo{title}{ImageNet Classification with
  Deep Convolutional Neural Networks}, in: \bibinfo{editor}{F.~Pereira},
  \bibinfo{editor}{C.~J.~C. Burges}, \bibinfo{editor}{L.~Bottou},
  \bibinfo{editor}{K.~Q. Weinberger} (Eds.), \bibinfo{booktitle}{Advances in
  Neural Information Processing Systems 25}, \bibinfo{publisher}{Curran
  Associates, Inc.}, \bibinfo{pages}{1097--1105},
  \urlprefix\url{http://papers.nips.cc/paper/4824-imagenet-classification-with-deep-convolutional-neural-networks.pdf},
  \bibinfo{year}{2012}.

\bibitem[{Girshick(2015)}]{girshick2015fast}
\bibinfo{author}{R.~Girshick}, \bibinfo{title}{Fast r-cnn}, in:
  \bibinfo{booktitle}{Proceedings of the IEEE international conference on
  computer vision}, \bibinfo{pages}{1440--1448}, \bibinfo{year}{2015}.

\bibitem[{Deng and Liu(2018)}]{deng2018deep}
\bibinfo{author}{L.~Deng}, \bibinfo{author}{Y.~Liu}, \bibinfo{title}{Deep
  Learning in Natural Language Processing}, \bibinfo{publisher}{Springer},
  \bibinfo{year}{2018}.

\bibitem[{Carrara et~al.(2018{\natexlab{a}})Carrara, Esuli, Fagni, Falchi, and
  Moreo~Fern{\'a}ndez}]{Carrara2018}
\bibinfo{author}{F.~Carrara}, \bibinfo{author}{A.~Esuli},
  \bibinfo{author}{T.~Fagni}, \bibinfo{author}{F.~Falchi},
  \bibinfo{author}{A.~Moreo~Fern{\'a}ndez}, \bibinfo{title}{Picture it in your
  mind: generating high level visual representations from textual
  descriptions}, \bibinfo{journal}{Information Retrieval Journal}
  \bibinfo{volume}{21}~(\bibinfo{number}{2})
  (\bibinfo{year}{2018}{\natexlab{a}}) \bibinfo{pages}{208--229}, ISSN
  \bibinfo{issn}{1573-7659}, 
  \urlprefix\url{https://doi.org/10.1007/s10791-017-9318-6}.

\bibitem[{Ortis et~al.(2019)Ortis, Farinella, and Battiato}]{OrtisFB19}
\bibinfo{author}{A.~Ortis}, \bibinfo{author}{G.~M. Farinella},
  \bibinfo{author}{S.~Battiato}, \bibinfo{title}{An Overview on Image Sentiment
  Analysis: Methods, Datasets and Current Challenges}, in:
  \bibinfo{booktitle}{Proceedings of the 16th International Joint Conference on
  e-Business and Telecommunications, {ICETE} 2019 - Volume 1: DCNET, ICE-B,
  OPTICS, {SIGMAP} and WINSYS, Prague, Czech Republic, July 26-28, 2019.},
  \bibinfo{pages}{296--306}, 
  \urlprefix\url{https://doi.org/10.5220/0007909602900300},
  \bibinfo{year}{2019}.

\bibitem[{Biggio et~al.(2013)Biggio, Corona, Maiorca, Nelson, {\v{S}}rndi{\'c},
  Laskov, Giacinto, and Roli}]{biggio2013evasion}
\bibinfo{author}{B.~Biggio}, \bibinfo{author}{I.~Corona},
  \bibinfo{author}{D.~Maiorca}, \bibinfo{author}{B.~Nelson},
  \bibinfo{author}{N.~{\v{S}}rndi{\'c}}, \bibinfo{author}{P.~Laskov},
  \bibinfo{author}{G.~Giacinto}, \bibinfo{author}{F.~Roli},
  \bibinfo{title}{Evasion attacks against machine learning at test time}, in:
  \bibinfo{booktitle}{Joint European conference on machine learning and
  knowledge discovery in databases}, \bibinfo{organization}{Springer},
  \bibinfo{pages}{387--402}, \bibinfo{year}{2013}.

\bibitem[{Szegedy et~al.(2013)Szegedy, Zaremba, Sutskever, Bruna, Erhan,
  Goodfellow, and Fergus}]{szegedy2013intriguing}
\bibinfo{author}{C.~Szegedy}, \bibinfo{author}{W.~Zaremba},
  \bibinfo{author}{I.~Sutskever}, \bibinfo{author}{J.~Bruna},
  \bibinfo{author}{D.~Erhan}, \bibinfo{author}{I.~Goodfellow},
  \bibinfo{author}{R.~Fergus}, \bibinfo{title}{Intriguing properties of neural
  networks}, \bibinfo{journal}{arXiv preprint arXiv:1312.6199} .

\bibitem[{Sundararajan and Woodard(2018)}]{sundararajan2018deep}
\bibinfo{author}{K.~Sundararajan}, \bibinfo{author}{D.~L. Woodard},
  \bibinfo{title}{Deep learning for biometrics: a survey},
  \bibinfo{journal}{ACM Computing Surveys (CSUR)}
  \bibinfo{volume}{51}~(\bibinfo{number}{3}) (\bibinfo{year}{2018})
  \bibinfo{pages}{65}.

\bibitem[{Cao et~al.(2018)Cao, Shen, Xie, Parkhi, and
  Zisserman}]{cao2018vggface2}
\bibinfo{author}{Q.~Cao}, \bibinfo{author}{L.~Shen}, \bibinfo{author}{W.~Xie},
  \bibinfo{author}{O.~M. Parkhi}, \bibinfo{author}{A.~Zisserman},
  \bibinfo{title}{Vggface2: A dataset for recognising faces across pose and
  age}, in: \bibinfo{booktitle}{2018 13th IEEE International Conference on
  Automatic Face \& Gesture Recognition (FG 2018)},
  \bibinfo{organization}{IEEE}, \bibinfo{pages}{67--74}, \bibinfo{year}{2018}.

\bibitem[{Amato et~al.(2018)Amato, Carrara, Falchi, Gennaro, and
  Vairo}]{amato2018facial}
\bibinfo{author}{G.~Amato}, \bibinfo{author}{F.~Carrara},
  \bibinfo{author}{F.~Falchi}, \bibinfo{author}{C.~Gennaro},
  \bibinfo{author}{C.~Vairo}, \bibinfo{title}{Facial-based Intrusion Detection
  System with Deep Learning in Embedded Devices}, in:
  \bibinfo{booktitle}{Proceedings of the 2018 International Conference on
  Sensors, Signal and Image Processing}, \bibinfo{organization}{ACM},
  \bibinfo{pages}{64--68}, \bibinfo{year}{2018}.

\bibitem[{Liu et~al.(2017)Liu, Wen, Yu, Li, Raj, and Song}]{liu2017sphereface}
\bibinfo{author}{W.~Liu}, \bibinfo{author}{Y.~Wen}, \bibinfo{author}{Z.~Yu},
  \bibinfo{author}{M.~Li}, \bibinfo{author}{B.~Raj}, \bibinfo{author}{L.~Song},
  \bibinfo{title}{Sphereface: Deep hypersphere embedding for face recognition},
  in: \bibinfo{booktitle}{Proceedings of the IEEE conference on computer vision
  and pattern recognition}, \bibinfo{pages}{212--220}, \bibinfo{year}{2017}.

\bibitem[{Feldstein(2019)}]{feldstein2019global}
\bibinfo{author}{S.~Feldstein}, \bibinfo{title}{The Global Expansion of AI
  Surveillance}, \bibinfo{type}{Working Paper}, \bibinfo{institution}{Carnegie
  Endowment for International Peace}, \bibinfo{address}{1779 Massachusetts
  Avenue NW, Washington, DC 20036},
  \urlprefix\url{https://carnegieendowment.org/files/WP-Feldstein-AISurveillance_final1.pdf},
  \bibinfo{year}{2019}.

\bibitem[{Dong et~al.(2019)Dong, Su, Wu, Li, Liu, Zhang, and
  Zhu}]{dong2019efficient}
\bibinfo{author}{Y.~Dong}, \bibinfo{author}{H.~Su}, \bibinfo{author}{B.~Wu},
  \bibinfo{author}{Z.~Li}, \bibinfo{author}{W.~Liu},
  \bibinfo{author}{T.~Zhang}, \bibinfo{author}{J.~Zhu},
  \bibinfo{title}{Efficient Decision-based Black-box Adversarial Attacks on
  Face Recognition}, in: \bibinfo{booktitle}{Proceedings of the IEEE Conference
  on Computer Vision and Pattern Recognition}, \bibinfo{pages}{7714--7722},
  \bibinfo{year}{2019}.

\bibitem[{Song et~al.(2018)Song, Wu, and Yang}]{song2018attacks}
\bibinfo{author}{Q.~Song}, \bibinfo{author}{Y.~Wu}, \bibinfo{author}{L.~Yang},
  \bibinfo{title}{Attacks on State-of-the-Art Face Recognition using
  Attentional Adversarial Attack Generative Network}, \bibinfo{journal}{arXiv
  preprint arXiv:1811.12026} .

\bibitem[{Sharif et~al.(2016)Sharif, Bhagavatula, Bauer, and
  Reiter}]{sharif2016accessorize}
\bibinfo{author}{M.~Sharif}, \bibinfo{author}{S.~Bhagavatula},
  \bibinfo{author}{L.~Bauer}, \bibinfo{author}{M.~K. Reiter},
  \bibinfo{title}{Accessorize to a crime: Real and stealthy attacks on
  state-of-the-art face recognition}, in: \bibinfo{booktitle}{Proceedings of
  the 2016 ACM SIGSAC Conference on Computer and Communications Security},
  \bibinfo{organization}{ACM}, \bibinfo{pages}{1528--1540},
  \bibinfo{year}{2016}.

\bibitem[{Kurakin et~al.(2016{\natexlab{a}})Kurakin, Goodfellow, and
  Bengio}]{kurakin2016adversarialphysworld}
\bibinfo{author}{A.~Kurakin}, \bibinfo{author}{I.~Goodfellow},
  \bibinfo{author}{S.~Bengio}, \bibinfo{title}{Adversarial examples in the
  physical world}, \bibinfo{journal}{arXiv preprint arXiv:1607.02533} .

\bibitem[{Li and Li(2017)}]{li2017adversarial}
\bibinfo{author}{X.~Li}, \bibinfo{author}{F.~Li}, \bibinfo{title}{Adversarial
  Examples Detection in Deep Networks with Convolutional Filter Statistics.},
  in: \bibinfo{booktitle}{ICCV}, \bibinfo{pages}{5775--5783},
  \bibinfo{year}{2017}.

\bibitem[{Liao et~al.(2018)Liao, Liang, Dong, Pang, Hu, and
  Zhu}]{liao2018defense}
\bibinfo{author}{F.~Liao}, \bibinfo{author}{M.~Liang},
  \bibinfo{author}{Y.~Dong}, \bibinfo{author}{T.~Pang},
  \bibinfo{author}{X.~Hu}, \bibinfo{author}{J.~Zhu}, \bibinfo{title}{Defense
  against adversarial attacks using high-level representation guided denoiser},
  in: \bibinfo{booktitle}{Proceedings of the IEEE Conference on Computer Vision
  and Pattern Recognition}, \bibinfo{pages}{1778--1787}, \bibinfo{year}{2018}.

\bibitem[{Kurakin et~al.(2016{\natexlab{b}})Kurakin, Goodfellow, and
  Bengio}]{kurakin2016adversarial}
\bibinfo{author}{A.~Kurakin}, \bibinfo{author}{I.~Goodfellow},
  \bibinfo{author}{S.~Bengio}, \bibinfo{title}{Adversarial examples in the
  physical world}, \bibinfo{journal}{arXiv preprint arXiv:1607.02533} .

\bibitem[{Papernot et~al.(2016)Papernot, McDaniel, Wu, Jha, and
  Swami}]{papernot2016distillation}
\bibinfo{author}{N.~Papernot}, \bibinfo{author}{P.~McDaniel},
  \bibinfo{author}{X.~Wu}, \bibinfo{author}{S.~Jha},
  \bibinfo{author}{A.~Swami}, \bibinfo{title}{Distillation as a defense to
  adversarial perturbations against deep neural networks}, in:
  \bibinfo{booktitle}{2016 IEEE Symposium on Security and Privacy (SP)},
  \bibinfo{organization}{IEEE}, \bibinfo{pages}{582--597},
  \bibinfo{year}{2016}.

\bibitem[{Gong et~al.(2017)Gong, Wang, and Ku}]{gong2017adversarial}
\bibinfo{author}{Z.~Gong}, \bibinfo{author}{W.~Wang}, \bibinfo{author}{W.-S.
  Ku}, \bibinfo{title}{Adversarial and clean data are not twins},
  \bibinfo{journal}{arXiv preprint arXiv:1704.04960} .

\bibitem[{Grosse et~al.(2017)Grosse, Manoharan, Papernot, Backes, and
  McDaniel}]{grosse2017statistical}
\bibinfo{author}{K.~Grosse}, \bibinfo{author}{P.~Manoharan},
  \bibinfo{author}{N.~Papernot}, \bibinfo{author}{M.~Backes},
  \bibinfo{author}{P.~McDaniel}, \bibinfo{title}{On the (statistical) detection
  of adversarial examples}, \bibinfo{journal}{arXiv preprint arXiv:1702.06280}
  .

\bibitem[{Amirian et~al.(2018)Amirian, Schwenker, and
  Stadelmann}]{amirian2018trace}
\bibinfo{author}{M.~Amirian}, \bibinfo{author}{F.~Schwenker},
  \bibinfo{author}{T.~Stadelmann}, \bibinfo{title}{Trace and detect adversarial
  attacks on CNNs using feature response maps}, in: \bibinfo{booktitle}{8th
  IAPR TC3 Workshop on Artificial Neural Networks in Pattern Recognition
  (ANNPR), Siena, Italy, September 19--21, 2018}, \bibinfo{organization}{IAPR},
  \bibinfo{year}{2018}.

\bibitem[{Metzen et~al.(2017)Metzen, Genewein, Fischer, and
  Bischoff}]{metzen2017detecting}
\bibinfo{author}{J.~H. Metzen}, \bibinfo{author}{T.~Genewein},
  \bibinfo{author}{V.~Fischer}, \bibinfo{author}{B.~Bischoff},
  \bibinfo{title}{On detecting adversarial perturbations},
  \bibinfo{journal}{arXiv preprint arXiv:1702.04267} .

\bibitem[{Carlini and Wagner(2017{\natexlab{a}})}]{carlini2017adversarial}
\bibinfo{author}{N.~Carlini}, \bibinfo{author}{D.~Wagner},
  \bibinfo{title}{Adversarial examples are not easily detected: Bypassing ten
  detection methods}, in: \bibinfo{booktitle}{Proceedings of the 10th ACM
  Workshop on Artificial Intelligence and Security},
  \bibinfo{organization}{ACM}, \bibinfo{pages}{3--14},
  \bibinfo{year}{2017}{\natexlab{a}}.

\bibitem[{Carrara et~al.(2019)Carrara, Falchi, Caldelli, Amato, and
  Becarelli}]{carrara2019adversarial}
\bibinfo{author}{F.~Carrara}, \bibinfo{author}{F.~Falchi},
  \bibinfo{author}{R.~Caldelli}, \bibinfo{author}{G.~Amato},
  \bibinfo{author}{R.~Becarelli}, \bibinfo{title}{Adversarial image detection
  in deep neural networks}, \bibinfo{journal}{Multimedia Tools and
  Applications} \bibinfo{volume}{78}~(\bibinfo{number}{3})
  (\bibinfo{year}{2019}) \bibinfo{pages}{2815--2835}.

\bibitem[{Papernot and McDaniel(2018)}]{papernot2018deep}
\bibinfo{author}{N.~Papernot}, \bibinfo{author}{P.~McDaniel},
  \bibinfo{title}{Deep k-nearest neighbors: Towards confident, interpretable
  and robust deep learning}, \bibinfo{journal}{arXiv preprint arXiv:1803.04765}
  .

\bibitem[{Sitawarin and Wagner(2019)}]{sitawarin2019robustness}
\bibinfo{author}{C.~Sitawarin}, \bibinfo{author}{D.~Wagner}, \bibinfo{title}{On
  the Robustness of Deep K-Nearest Neighbors}, \bibinfo{journal}{arXiv preprint
  arXiv:1903.08333} .

\bibitem[{Kendall and Gal(2017)}]{kendall2017uncertainties}
\bibinfo{author}{A.~Kendall}, \bibinfo{author}{Y.~Gal}, \bibinfo{title}{What
  uncertainties do we need in bayesian deep learning for computer vision?}, in:
  \bibinfo{booktitle}{Advances in neural information processing systems},
  \bibinfo{pages}{5574--5584}, \bibinfo{year}{2017}.

\bibitem[{Carrara et~al.(2018{\natexlab{b}})Carrara, Becarelli, Caldelli,
  Falchi, and Amato}]{carrara2018adversarial}
\bibinfo{author}{F.~Carrara}, \bibinfo{author}{R.~Becarelli},
  \bibinfo{author}{R.~Caldelli}, \bibinfo{author}{F.~Falchi},
  \bibinfo{author}{G.~Amato}, \bibinfo{title}{Adversarial examples detection in
  features distance spaces}, in: \bibinfo{booktitle}{Proceedings of the
  European Conference on Computer Vision (ECCV)}, \bibinfo{pages}{0--0},
  \bibinfo{year}{2018}{\natexlab{b}}.

\bibitem[{Whitelam et~al.(2017)Whitelam, Taborsky, Blanton, Maze, Adams,
  Miller, Kalka, Jain, Duncan, Allen et~al.}]{whitelam2017iarpa}
\bibinfo{author}{C.~Whitelam}, \bibinfo{author}{E.~Taborsky},
  \bibinfo{author}{A.~Blanton}, \bibinfo{author}{B.~Maze},
  \bibinfo{author}{J.~Adams}, \bibinfo{author}{T.~Miller},
  \bibinfo{author}{N.~Kalka}, \bibinfo{author}{A.~K. Jain},
  \bibinfo{author}{J.~A. Duncan}, \bibinfo{author}{K.~Allen}, et~al.,
  \bibinfo{title}{Iarpa janus benchmark-b face dataset}, in:
  \bibinfo{booktitle}{Proceedings of the IEEE Conference on Computer Vision and
  Pattern Recognition Workshops}, \bibinfo{pages}{90--98},
  \bibinfo{year}{2017}.

\bibitem[{Maze et~al.(2018)Maze, Adams, Duncan, Kalka, Miller, Otto, Jain,
  Niggel, Anderson, Cheney et~al.}]{maze2018iarpa}
\bibinfo{author}{B.~Maze}, \bibinfo{author}{J.~Adams}, \bibinfo{author}{J.~A.
  Duncan}, \bibinfo{author}{N.~Kalka}, \bibinfo{author}{T.~Miller},
  \bibinfo{author}{C.~Otto}, \bibinfo{author}{A.~K. Jain},
  \bibinfo{author}{W.~T. Niggel}, \bibinfo{author}{J.~Anderson},
  \bibinfo{author}{J.~Cheney}, et~al., \bibinfo{title}{IARPA janus benchmark-c:
  Face dataset and protocol}, in: \bibinfo{booktitle}{2018 International
  Conference on Biometrics (ICB)}, \bibinfo{organization}{IEEE},
  \bibinfo{pages}{158--165}, \bibinfo{year}{2018}.

\bibitem[{Goodfellow et~al.(2014)Goodfellow, Shlens, and
  Szegedy}]{goodfellow2014explaining}
\bibinfo{author}{I.~J. Goodfellow}, \bibinfo{author}{J.~Shlens},
  \bibinfo{author}{C.~Szegedy}, \bibinfo{title}{Explaining and harnessing
  adversarial examples (2014)}, \bibinfo{journal}{arXiv preprint
  arXiv:1412.6572} .

\bibitem[{Dong et~al.(2018)Dong, Liao, Pang, Su, Zhu, Hu, and
  Li}]{dong2018boosting}
\bibinfo{author}{Y.~Dong}, \bibinfo{author}{F.~Liao},
  \bibinfo{author}{T.~Pang}, \bibinfo{author}{H.~Su}, \bibinfo{author}{J.~Zhu},
  \bibinfo{author}{X.~Hu}, \bibinfo{author}{J.~Li}, \bibinfo{title}{Boosting
  adversarial attacks with momentum}, \bibinfo{journal}{arXiv preprint} .

\bibitem[{Carlini and Wagner(2017{\natexlab{b}})}]{carlini2017towards}
\bibinfo{author}{N.~Carlini}, \bibinfo{author}{D.~Wagner},
  \bibinfo{title}{Towards evaluating the robustness of neural networks}, in:
  \bibinfo{booktitle}{2017 IEEE Symposium on Security and Privacy (SP)},
  \bibinfo{organization}{IEEE}, \bibinfo{pages}{39--57},
  \bibinfo{year}{2017}{\natexlab{b}}.

\bibitem[{Turk and Pentland(1991)}]{turk1991face}
\bibinfo{author}{M.~A. Turk}, \bibinfo{author}{A.~P. Pentland},
  \bibinfo{title}{Face recognition using eigenfaces}, in:
  \bibinfo{booktitle}{Proceedings. 1991 IEEE Computer Society Conference on
  Computer Vision and Pattern Recognition}, \bibinfo{organization}{IEEE},
  \bibinfo{pages}{586--591}, \bibinfo{year}{1991}.

\bibitem[{Wang and Deng(2018)}]{wang2018deep}
\bibinfo{author}{M.~Wang}, \bibinfo{author}{W.~Deng}, \bibinfo{title}{Deep face
  recognition: A survey}, \bibinfo{journal}{arXiv preprint arXiv:1804.06655} .

\bibitem[{Massoli et~al.(2019)Massoli, Amato, Falchi, Gennaro, and
  Vairo}]{massoli2019improving}
\bibinfo{author}{F.~V. Massoli}, \bibinfo{author}{G.~Amato},
  \bibinfo{author}{F.~Falchi}, \bibinfo{author}{C.~Gennaro},
  \bibinfo{author}{C.~Vairo}, \bibinfo{title}{Improving Multi-scale Face
  Recognition Using VGGFace2}, in: \bibinfo{booktitle}{International Conference
  on Image Analysis and Processing}, \bibinfo{organization}{Springer},
  \bibinfo{pages}{21--29}, \bibinfo{year}{2019}.

\bibitem[{Qiu et~al.(2019)Qiu, Xiao, Yang, Yan, Lee, and
  Li}]{qiu2019semanticadv}
\bibinfo{author}{H.~Qiu}, \bibinfo{author}{C.~Xiao}, \bibinfo{author}{L.~Yang},
  \bibinfo{author}{X.~Yan}, \bibinfo{author}{H.~Lee}, \bibinfo{author}{B.~Li},
  \bibinfo{title}{SemanticAdv: Generating Adversarial Examples via
  Attribute-conditional Image Editing}, \bibinfo{journal}{arXiv preprint
  arXiv:1906.07927} .

\bibitem[{Kakizaki and Yoshida(2019)}]{kakizaki2019adversarial}
\bibinfo{author}{K.~Kakizaki}, \bibinfo{author}{K.~Yoshida},
  \bibinfo{title}{Adversarial Image Translation: Unrestricted Adversarial
  Examples in Face Recognition Systems}, \bibinfo{year}{2019}.

\bibitem[{Pautov et~al.(2019)Pautov, Melnikov, Kaziakhmedov, Kireev, and
  Petiushko}]{pautov2019adversarial}
\bibinfo{author}{M.~Pautov}, \bibinfo{author}{G.~Melnikov},
  \bibinfo{author}{E.~Kaziakhmedov}, \bibinfo{author}{K.~Kireev},
  \bibinfo{author}{A.~Petiushko}, \bibinfo{title}{On adversarial patches:
  real-world attack on ArcFace-100 face recognition system},
  \bibinfo{journal}{arXiv preprint arXiv:1910.07067} .

\bibitem[{Huang et~al.(2015)Huang, Xu, Schuurmans, and
  Szepesv{\'a}ri}]{huang1511learning}
\bibinfo{author}{R.~Huang}, \bibinfo{author}{B.~Xu},
  \bibinfo{author}{D.~Schuurmans}, \bibinfo{author}{C.~Szepesv{\'a}ri},
  \bibinfo{title}{Learning with a strong adversary. arXiv 2015},
  \bibinfo{journal}{arXiv preprint arXiv:1511.03034} .

\bibitem[{Papernot et~al.(2015)Papernot, McDaniel, Wu, Jha, and
  Swami}]{papernot2015distillation}
\bibinfo{author}{N.~Papernot}, \bibinfo{author}{P.~McDaniel},
  \bibinfo{author}{X.~Wu}, \bibinfo{author}{S.~Jha},
  \bibinfo{author}{A.~Swami}, \bibinfo{title}{Distillation as a defense to
  adversarial perturbations against deep neural networks},
  \bibinfo{journal}{arXiv preprint arXiv:1511.04508} .

\bibitem[{Goswami et~al.(2019)Goswami, Agarwal, Ratha, Singh, and
  Vatsa}]{goswami2019detecting}
\bibinfo{author}{G.~Goswami}, \bibinfo{author}{A.~Agarwal},
  \bibinfo{author}{N.~Ratha}, \bibinfo{author}{R.~Singh},
  \bibinfo{author}{M.~Vatsa}, \bibinfo{title}{Detecting and mitigating
  adversarial perturbations for robust face recognition},
  \bibinfo{journal}{International Journal of Computer Vision}
  \bibinfo{volume}{127}~(\bibinfo{number}{6-7}) (\bibinfo{year}{2019})
  \bibinfo{pages}{719--742}.

\bibitem[{Sabour et~al.(2015)Sabour, Cao, Faghri, and
  Fleet}]{sabour2015adversarial}
\bibinfo{author}{S.~Sabour}, \bibinfo{author}{Y.~Cao},
  \bibinfo{author}{F.~Faghri}, \bibinfo{author}{D.~J. Fleet},
  \bibinfo{title}{Adversarial manipulation of deep representations},
  \bibinfo{journal}{arXiv preprint arXiv:1511.05122} .

\bibitem[{Kingma and Ba(2014)}]{kingma2014adam}
\bibinfo{author}{D.~P. Kingma}, \bibinfo{author}{J.~Ba}, \bibinfo{title}{Adam:
  A method for stochastic optimization}, \bibinfo{journal}{arXiv preprint
  arXiv:1412.6980} .

\end{thebibliography}
\end{document}